\documentclass[letterpaper]{article} %

\usepackage{aaai23}  %
\usepackage[]{aaai23_extensions}
\usepackage{flushend}

\newif\ifseparateappendix

\separateappendixtrue

\ifseparateappendix
\usepackage{xr-latexmk}
\myexternaldocument{aaai23-appendix}
\else
\usepackage[page]{appendix}
\fi

\usepackage{times}  %
\usepackage{helvet}  %
\usepackage{courier}  %
\usepackage[hyphens]{url}  %
\usepackage{graphicx} %
\usepackage{natbib}  %
\usepackage{caption} %
\usepackage{algorithm}
\usepackage{algorithmic}

\usepackage{newfloat}
\usepackage{listings}
\DeclareCaptionStyle{ruled}{labelfont=normalfont,labelsep=colon,strut=off} %
\floatstyle{ruled}
\newfloat{listing}{tb}{lst}{}
\floatname{listing}{Listing}
\usepackage{gltl_macros}
\usepackage{paper_macros}
\usepackage[]{marginalia}
\usepackage[noframe]{showframe}

\usepackage{enumitem}
\setlist{nolistsep}
\setlist[itemize]{noitemsep}
\usetikzlibrary{external}

\title{Computably Continuous Reinforcement-Learning Objectives are PAC-learnable}

\author {
    Cambridge Yang,\textsuperscript{\rm 1}
    Michael Littman, \textsuperscript{\rm 2}
    Michael Carbin \textsuperscript{\rm 1}
}
\affiliations {
    \textsuperscript{\rm 1} MIT\\
    \textsuperscript{\rm 2} Brown University \\
    camyang@csail.mit.edu, 
    mlittman@cs.brown.edu, 
    mcarbin@csail.mit.edu
}

\makeatletter\@input{aaai23-aux.tex}\makeatother
\makeatletter\@input{aaai23-appendix-aux.tex}\makeatother

\begin{document}

\maketitle

\begin{abstract}
In reinforcement learning, the classic objectives of maximizing discounted and finite-horizon cumulative rewards are PAC-learnable: There are algorithms that learn a near-optimal policy with high probability using a finite amount of samples and computation. 
In recent years, researchers have introduced objectives and corresponding reinforcement-learning algorithms beyond the classic cumulative rewards, such as objectives specified as linear temporal logic formulas. 
However, questions about the PAC-learnability of these new objectives have remained open.

This work demonstrates the PAC-learnability of general reinforcement-learning objectives through sufficient conditions for PAC-learnability in two analysis settings. 
In particular, for the analysis that considers only sample complexity, we prove that if an objective given as an oracle is uniformly continuous, then it is PAC-learnable.
Further, for the analysis that considers computational complexity, we prove that if an objective is computable, then it is PAC-learnable. 
In other words, if a procedure computes successive approximations of the objective's value, then the objective is PAC-learnable.

We give three applications of our condition on objectives from the literature with previously unknown PAC-learnability and prove that these objectives are PAC-learnable.  
Overall, our result helps verify existing objectives' PAC-learnability. 
Also, as some studied objectives that are not uniformly continuous have been shown to be not PAC-learnable, our results could guide the design of new PAC-learnable objectives.
\end{abstract}

\newif\ifshorterpaper

\shorterpapertrue

\section{Introduction}

In reinforcement learning, we situate an agent in an environment with unknown dynamics. 
The agent acts in the environment by executing its current policy. Executing a policy in an environment induces an
infinite-length path of states and actions. We specify an {\em objective}, 
a function that maps each possible infinite-length path to a real number---a score---for that path. 
Moreover, we request the agent learn a good policy that nearly maximizes the expected score over the distribution of paths induced by the environment and the policy.

\paragraph{PAC-learnability of Objectives}
The classic \reinforcementlearning objectives include infinite-horizon discounted cumulative rewards and finite-horizon cumulative rewards. 
These objectives are well-studied and have a desirable property: 
There are \reinforcementlearning algorithms that learn a near-optimal policy with high probability with number of samples depending only on the parameters known by the algorithm. 
We call these algorithms probably approximately correct (PAC), and these objectives {\em PAC-learnable} under reinforcement learning.

PAC-learnability is essential. 
Specifically, we aim to specify an objective and let the agent learn a good policy on its own. 
Thus, we necessarily need some form of assurance of how close to optimal the learned policy is. If an objective is not PAC-learnable, then the hope of ensuring learning a near-optimal policy is lost, and the objective is effectively intractable to learn under reinforcement learning.

\paragraph{General Objectives}
In recent years, researchers have introduced various objectives beyond the two classic rewards objectives~\citep{smcbltl,dorsa14,gltl17,hasanbeig2019reinforcement,omegaregularrl19,rewardmachine1,ltlf-rl,osbert,bozkurt2020control,rdp21}. 
For example:
\begin{itemize}[wide]
\item \citet{rewardmachine1} introduced the reward machine objective. A reward machine augments classic rewards with an automaton that makes the rewards history dependent.
\item \citet{bozkurt2020control} introduced an objective based on limit deterministic Buchi Automaton (LDBA).\footnote{We summarize works along the same line in \Cref{sec:summary-rl-automaton-objectives}.}
The objective features history-dependent discount factors, history-dependent rewards, and an augmented action space. 
\end{itemize}
Researchers introduced \reinforcementlearning algorithms for these objectives and showed that they empirically learn well-behaving policies with finitely many samples.

\paragraph{PAC-learnability of General Objectives}

Despite the advances on empirical algorithms for these objectives, not all objectives are PAC-learnable:
Recent work \citep{ltlnonpac22} proved that infinite-horizon linear temporal logic objectives, a class of general objectives, are not PAC-learnable.
Therefore, to the end of having assurance on the outcomes of learning, we desire to understand the PAC\Hyphdash learnability of general objectives.

Some previous work~\citep{pacmdpsmcconfidenceinterval,smcbltl,ufukpacmdpltl14,rdp21} address the PAC-learnability of particular objectives. 
However, these analyses give \reinforcementlearning algorithms for particular objectives and do not generalize to other objectives.
Previous work \citep{alur2021framework} gave a framework of reductions between objectives whose flavor of generality is most similar to our work; however, they did not give a condition for when an objective is PAC-learnable.
To our knowledge, the PAC-learnability of the objectives in  \citet{dorsa14,gltl17,hasanbeig2019reinforcement,omegaregularrl19,rewardmachine1,osbert,bozkurt2020control} are not known. 

Relevant to model-based reinforcement learning, \citet{bazille20} showed that it is impossible to learn the transitions of a Markov chain such that the learned and true models agree on all first-order behaviors. 
However, this result does not apply to the general reinforcement-learning setting.

We thus raise a research question: {\em When is a \reinforcementlearning objective PAC-learnable?}

\paragraph{Our Approach}
We address the question by giving sufficient conditions for PAC-learnability.
Specifically, we analyze PAC-learnability in both the information-theoretic setting, that only considers sample complexity, and the computation\Hyphdash theoretic setting, that also considers computability.
We prove that, in the information-theoretic setting (resp. computation-theoretic setting), an objective is PAC\Hyphdash learnable if it is uniformly continuous (resp. computable).
These conditions simplify proving objectives' PAC-learnability.
In particular, our conditions avoid constraints on environments, policies, or \reinforcementlearning algorithms but only require reasoning about the objective itself.

We give example applications of our conditions to three objectives in the literature whose PAC-learnability was previously unknown and prove that they are PAC-learnable.

\paragraph{Contributions}
We make the following contributions about \reinforcementlearning objectives:
\begin{itemize}[wide]
\item In the information-theoretic setting, we prove that a uniformly continuous objective is PAC-learnable.
\item In the computation-theoretic setting, we prove that a computable objective is PAC-learnable.
\item We apply our theorem to three objectives \citep{rewardmachine1,bozkurt2020control,gltl17} from the literature whose PAC-learnability was previously unknown and show that they are PAC-learnable.
\end{itemize}
Our result makes checking the PAC-learnability of existing objectives easier. 
It also potentially guides the design of new PAC-learnable objectives.

\section{Reinforcement Learning and Objectives}

This section reviews reinforcement learning and defines general objectives and their learnability.

\subsection{Markov Processes}

A {\em Markov decision process} (MDP) is a tuple $\mathcal{M} = (S,A,P,s_0)$, where $S$ and $A$ are finite sets of states and actions, $\functiontype[P]{S, A}{\distributionon{S}}$ is a transition probability function that maps a current state and an action to a distribution over next states, and $s_0 \in S$ is an initial state. The MDP is sometimes referred to as the \emph{environment MDP} to distinguish it from any specific objective.

A {\em policy} for an MDP is a function $\functiontype[\pi]{\left(S \times A\right)^{*}, S}{\distributionon{A}}$ mapping a history of states and actions to a distribution over actions. \footnote{\label{seq-notation} $X^*$ denotes all finite-length sequences of the elements of $X$. $\infseq{X}$ denotes all infinite-length sequences of the elements of $X$.}
An MDP and a policy induce a {\em discrete-time Markov chain} (DTMC). 
A DTMC is a tuple $\mathcal{D} = (S, P, s_0)$, where $S$ is the set of states, $\functiontype[P]{S}{\distributionon{S}}$ is a transition-probability function mapping states to distributions over next states, and $s_0 \in S$ is an initial state.
The DTMC induces a probability space over the infinite-length sequences $w \in \infseq{S}$.\footnotemark[\getrefnumber{seq-notation}]

\subsection{Objective}

\paragraph{Environment-specific Objective.}
An \begingroup\em environment\Hyphdash specific objective \endgroup for an MDP $(S, A, P, s_0)$ is a measurable function $\functiontype[\kappa]{\infseq{\left(S\times A\right)}}{\reals}$. 
We say such an objective is environment-specific since it is associated with MDPs with a fixed set of states and actions.

The {\em value} of an environment-specific objective for an MDP $\mathcal{M}$ and a policy $\pi$ is the expectation of the objective under the probability space of the DTMC $\mathcal{D}$ induced by $\mathcal{M}$ and $\pi$:
$
\mdpvaluefunc{\pi}{\mathcal{M}}{\kappa} = \expectation*{w \sim \mathcal{D}}{\kappa\left(w\right)}.
$
We consider only bounded objectives to ensure that the expectation exists and is finite.
The {\em optimal value} is the supremum of the values achievable by all policies: $\mdpvaluefunc{*}{\mathcal{M}}{\kappa} = \sup_{\pi} \mdpvaluefunc{\pi}{\mathcal{M}}{\kappa}$. 
A policy $\pi$ is $\epsilon$-optimal if its value is $\epsilon$-close to the optimal value: $\mdpvaluefunc{\pi}{\mathcal{M}}{\kappa} \ge \mdpvaluefunc{*}{\mathcal{M}}{\kappa} - \epsilon$.

\paragraph{Environment-generic Objective.}
An objective defined above is environment-specific because it is associated with a fixed set of states and actions.
However, we would also like to talk about objectives in a form decoupled from any MDP. 
For example, the discounted classical cumulative rewards objective is not bound to any particular reward function.
Further, the objective of ``reaching the goal state'' in a grid world environment is not bound to the size of the grid or the allowed actions.
Such decoupling is desirable as it allows one to specify objectives independent of environments.

To that end, we define {\em environment-generic} objectives.
The idea of such objectives is that a {\em labeling function} interfaces between the environment and the environment-generic objective.
The definition decouples an environment-generic objective from environments by requiring different labeling functions for different environments.

Formally, an {\em environment-generic objective} is a measurable function: $\functiontype[\xi]{\infseq{F}}{\reals}$, where $F$ is a set called {\em features}.
A {\em labeling function} maps the MDP's (current) states and actions to the features: $\functiontype[\mathcal{L}]{S, A}{F}$.
Composing $\xi$ and the element-wise application of $\mathcal{L}$ induces an environment-specific objective.
For example, the discounted cumulative rewards objective $\functiontype[\xi]{\infseq{\rationals}}{\reals}$ is $\xi(w) = \sum_{i=0}^{\infty} \gamma^i \cdot \windex{w}{i}$. For each MDP, the labeling function is a classical reward function $\functiontype[\mathcal{L}]{S, A}{\rationals}$. \footnote{For simplicity of analysis, we will let objective specifications use rationals instead of reals so that they admit a finite representation. Nonetheless, our analyses also generalize to objective specifications that contain reals. }

The value of $\xi$ for an MDP $\mathcal{M}$, a policy $\pi$, and a labeling function $\mathcal{L}$ is the value of the environment-specific objective induced by $\xi$, $\mathcal{M}$, $\pi$, and $\mathcal{L}$.

\subsection{Learning Model}

A reinforcement learning agent has access to a sampler of the MDP's transitions but does know the underlying probability values.
The agent learns in two phases: sampling and learning. 
In the sampling phase, the agent starts from the initial state and follows a sequence of decision steps to collect sampled environment transitions. 
At every step, the agent may
\begin{enumerate*}[(1)]
    \item act from the current state to sample the next state or
    \item reset to the initial state.
\end{enumerate*}
In the learning phase, it learns a policy from the collected sampled transitions.

Formally, a \reinforcementlearning algorithm is a tuple \begingroup\small$(\mathcal{A}^S, \mathcal{A}^L)$\endgroup, where \begingroup\small$\mathcal{A}^S$ 
\endgroup is a {\em sampling algorithm} that drives how the environment is sampled, and $\mathcal{A}^L$ is a learning algorithm that learns a policy from the samples obtained by the sampling algorithm.
Let \begingroup\small$A_{\texttt{reset}} = A \cup \{\texttt{reset}\}$ \endgroup be the set of actions with an additional \texttt{reset} operation,
the sampling algorithm \begingroup\small$\functiontype[\mathcal{A}^S]{\left(\left(S \times A_{\texttt{reset}}\right)^* \times S\right)}{A_{\texttt{reset}}}$ \endgroup maps from sampled transitions to the next operation.
The learning algorithm \begingroup\small$\functiontype[\mathcal{A}^L]{\left(\left(S \times A_{\texttt{reset}}\right)^* \times S\right)}{\left(\functiontype{\left(\left(S \times A\right)^* \times S\right)}{\Delta(A)}\right)}$ \endgroup maps from the sampled transitions to the learned policy.

\subsection{Learnability of Objectives}

A good \reinforcementlearning algorithm should learn the optimal policy that maximizes the given objective.
In particular, we let the algorithm seek a near-optimal policy with high probability.

\begin{definition}[PAC Algorithm for Environment-specific Objective]
Given an objective $\kappa$, a \reinforcementlearning algorithm  $(\mathcal{A}^\text{S}, \mathcal{A}^\text{L})$ is $\kappa$-PAC (probably approximately correct for objective $\kappa$) in an environment MDP $\mathcal{M}$ with $N$ samples if, with the sequence of transitions $T$ of length $N$ sampled using the sampling algorithm $\mathcal{A}^\text{S}$, the learning algorithm $\mathcal{A}^\text{L}$ outputs an $\epsilon$-optimal policy with probability at least $1-\delta$ for any given $\epsilon>0$ and $0 < \delta < 1$. That is:
\begingroup\small\begin{equation*}
\prob*{\rv{T} \sim \mdpsamplingproduct{\mathcal{M}}{\mathcal{A}^\text{S}}{N}}{\mdpvaluefunc{\mathcal{A}^\text{L}\left(T\right)}{\mathcal{M}}{\kappa} \ge \mdpvaluefunc{*}{\mathcal{M}}{\kappa} - \epsilon} \ge 1 - \delta.
\end{equation*}\endgroup
\end{definition}
We use \begingroup\small$ \rv{T}{\sim}\mdpsamplingproduct{\mathcal{M}}{\mathcal{A}^\text{S}}{N}$\endgroup to denote that the probability space is over the set of length-$N$ transition sequences sampled from the environment $\mathcal{M}$ using the sampling algorithm $\mathcal{A}^\text{S}$.
We will simply write $\prob{\rv{T}}{.}$ when it is clear from the context.

We will consider two settings: the information-theoretic setting that considers only sample complexity and the computation-theoretic setting that considers computability.

\begin{definition}[PAC-learnable Environment-specific Objective]
In the information-theoretic setting (resp. computation-theoretic setting), an environment\Hyphdash specific objective $\kappa$ is {\em $\kappa$-PAC-learnable} if there exists a function $\functiontype[C]{\reals, \reals, \naturals, \naturals}{\naturals}$ such that, for all consistent environment MDPs for $\kappa$ (i.e., the domain of $\kappa$ uses the same set of states and actions as the MDP), 
there exists a $\kappa$-PAC \reinforcementlearning algorithm with less than $C(\frac{1}{\epsilon}, \frac{1}{\delta}, |S|, |A|)$ samples (resp. computation steps).

\end{definition}
Our definition focuses on the core tractability issue. 
Failure to respect our definition implies that PAC\Hyphdash learning is not achievable with finitely many samples (in the information-theoretic setting) or not computable (in the computation-theoretic setting). 
To that end, we have set the parameters of $C$ to be the only quantities available to an algorithm under the standard assumptions of reinforcement learning. 
Specifically, since the transition dynamics are unknown, they are not parameters of $C$.
Moreover, while some variants of PAC-learnability require $C$ to be a polynomial to capture the notion of learning efficiency, we have dropped this requirement to focus on the core tractability issue.

We also define the PAC-learnability of environment\Hyphdash generic objectives, for both information- and computation\Hyphdash theoretic settings:
\begin{definition}[PAC-learnable Environment-generic Objective]
An environment-generic objective $\xi$ is {\em $\xi$-PAC-learnable} if for all labeling functions $\mathcal{L}$, the objective $\kappa$ induced by  $\xi$ and $\mathcal{L}$ is $\kappa$-PAC-learnable.
\end{definition}

Note that in the information-theoretic setting, we assume that the objectives $\kappa$ and $\xi$ are given as oracles: they take infinite-length inputs and return infinite-precision output with no computation overhead.

\subsection{Established PAC-Learnable Objectives}

The standard discounted cumulative rewards objective $\sum_{i=0}^{\infty} \gamma^i \windex{w}{i}$ and the finite-horizon cumulative rewards objective $\sum_{i=0}^{H} \windex{w}{i}$ are known to be PAC-learnable.
The folklore intuition is that these objectives ``effectively terminate'' in an expected finite-length horizon, and rewards farther out of the horizon diminish quickly. 
This paper formalizes this intuition by connecting it to the standard definition of the objective function's uniform continuity and computability.
Later in \Cref{sec:uniform_continuity_implies_pac}, we will prove that uniformly continuous and computable objectives are PAC-learnable.

\section{Example: The Reward Machine Objective}
\label{sec:reward_machine_example}

This section gives an example general objective: the simple reward machine objective  \citep{rewardmachine1}.
We will later use this objective as one of the examples to apply our core theorem and prove its PAC-learnability.

\paragraph{The Simple Reward Machine Specification}
Simple reward machines generalize from classic Markovian rewards to non-Markovian rewards.
In particular, a simple reward machine is a kind of deterministic finite automaton. 
Each automaton transition has a reward value and a tuple of truth values for a set of propositions about the environment's state.
The simple reward machine starts from an initial state. 
As the agent steps through the environment, a labeling function classifies the environment's current state to a tuple of truth values of a set of propositions. 
The simple reward machine then transits to the next states according to the tuple.
During each transition, the agent collects a scalar reward along the transition of the simple reward machine.
The overall objective is to maximize the $\gamma$-discounted sum of collected rewards.
The formal definition of a simple reward machine given by \citet{rewardmachine1} is:
\begin{definition}[Simple Reward Machine]
Given a finite set $\Pi$ called the propositions, a simple reward machine over $\Pi$ is a tuple $(U, \delta_u, \delta_r, u_0, \gamma)$, where $U$ is a finite set of states, $\functiontype[\delta_u]{U, 2^\Pi}{U}$ is a deterministic state transition function, $\functiontype[\delta_r]{U, U}{\rationals}$ is a deterministic reward function, $u_0$ is an initial state, and $\gamma \in \rationals$ is a discount factor.
\end{definition}

\begin{figure}[t]
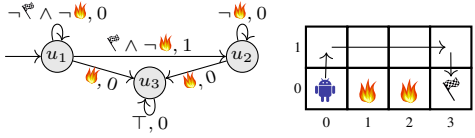

\tikzset{external/export next=false}
\centering
\begin{tikzpicture}[node distance=4.5em,on grid,auto]
\let\inputhidefromlatexpp\input

\tikzset{%
    in place/.style={
      auto=false,
      fill=white,
      inner sep=2pt,
    },
}

\newcommand{\statenodesize}{1.2em}

\tikzset{%
    every state/.style={
        fill={rgb:black,1;white,10},
        initial text=, 
        inner sep=0, 
        minimum size=\statenodesize,
        font={\small},
    },
    tl/.style={font=\small} %
}

\tikzset{%
    accstate/.style={
        state,
        pattern={Lines[angle=35,distance={4.5pt/sqrt(2)}]},
        pattern color=gray
    }
}

\tikzset{%
    rejstate/.style={state, fill=white}
}

\newcommand{\largedots}{$\textbf{\ldots}$}

\tikzset{policy/.style={color=gray}}
\definecolor{slipcolor}{HTML}{a3cef1}
\definecolor{trapcolor}{HTML}{eb4724}
\tikzset{
lava/.pic={code={\input{lava.tikz}}},
fire/.pic={code={\input{fire.tikz}}},
treasure/.pic={code={\input{treasure.tikz}}},
goal/.pic={code={\input{goal.tikz}}},
bot/.pic={code={\input{bot.tikz}}}
}

\tikzset{
  box/.style={rectangle,draw=black,thick,minimum size=1cm},
  slip/.style={box,fill=slipcolor},
  trap/.style={box,fill=trapcolor},
  wall/.style={box,fill=gray},
}

\begingroup
    \tikzstyle{every state}=[
        fill={rgb:black,1;white,10},
        initial text=, 
        inner sep=0, 
        minimum size=1.2em,
        font={\scriptsize},
    ]
    
    \tikzstyle{tl}=[style={font={\scriptsize}}] %

    \tikzstyle{every loop}=[
    style={
        looseness=1, 
        min distance=3mm,
        font={\scriptsize}
        }
    ]
    
    \node[state, initial left] (u1) {$u_1$};
    \node[state] (u2) [right =7em of u1]  {$u_2$};
    \node[state] (u3) [below right =1em and 3.5em of u1]  {$u_3$};

    \renewcommand{\goal}{\text{\begin{tikzpicture}[scale=0.4, transform shape, yshift=-10em]{\pic {goal};}\end{tikzpicture}}}
    \renewcommand{\lava}{\text{\begin{tikzpicture}[scale=0.4, transform shape, yshift=-10em]{\pic {lava};}\end{tikzpicture}}}

    \path[->]
    (u1) edge node [tl, above=-3pt, sloped] {$\goal \land \neg \lava, 1$} (u2)
    (u1) edge node [tl, below=-4pt, sloped] {$\lava,0$} (u3)
    (u2) edge node [tl, below=-4pt, sloped] {$\lava,0$} (u3)
    ;

    \path[->] 
    (u1) edge  [loop above]  node [tl, above=-3pt] {$\neg\goal\land \neg\lava,0$}  ()
    (u2) edge  [loop above]  node [tl, above=-3pt] {$\neg\lava, 0$}  ()
    (u3) edge  [loop below]  node [tl, below=-3pt] {$\top, 0$}  ()
;

\endgroup
\end{tikzpicture}
\tikzset{external/export next=false}
\centering
\begin{tikzpicture}[scale=0.55, transform shape, baseline={-2em}]
\let\inputhidefromlatexpp\input

\tikzset{policy/.style={color=gray}}
\definecolor{slipcolor}{HTML}{a3cef1}
\definecolor{trapcolor}{HTML}{eb4724}
\tikzset{
lava/.pic={code={\input{lava.tikz}}},
fire/.pic={code={\input{fire.tikz}}},
treasure/.pic={code={\input{treasure.tikz}}},
goal/.pic={code={\input{goal.tikz}}},
bot/.pic={code={\input{bot.tikz}}}
}

\tikzset{
  box/.style={rectangle,draw=black,thick,minimum size=1cm},
  slip/.style={box,fill=slipcolor},
  trap/.style={box,fill=trapcolor},
  wall/.style={box,fill=gray},
}

\foreach \x in {0,...,3}{
    \foreach \y in {0,...,1}
        \node[box] at (\x,\y){};
}
\foreach \x in {0,...,3}{
    \node at (\x, -0.8) {\x};
}
\foreach \y in {0,...,1}{
    \node at (-0.7, \y) {\y};
}

\node (N1) at (0, 1) {};
\node (N2) at (3, 1) {};

\pic [local bounding box=P] at (0, 0) {bot};
\pic [local bounding box=L1] at (1, 0) {lava};
\pic [local bounding box=L2] at (2, 0) {lava};
\pic [local bounding box=G] at (3, 0) {goal};

\draw[->]
    (P) edge (N1);

\draw[->]
    (N1) edge (N2);

\draw[->]
    (N2) edge (G);

\end{tikzpicture}
\caption{Left: simple reward machine. Right: environment.}
\label{fig:reward-machine}
\end{figure}
\Cref{fig:reward-machine} shows an example simple reward machine and an accompanying grid environment.
The states of the simple reward machine are $\{u_1, u_2, u_3\}$. 
\let\inputhidefromlatexpp\input
\tikzset{
lava/.pic={code={\definecolor{ce73900}{RGB}{231,57,0}
\definecolor{ce78c00}{RGB}{231,140,0}
\definecolor{ce7e100}{RGB}{231,225,0}
\definecolor{cffff9e}{RGB}{255,255,158}

\begin{scope}[shift={(-0.21,0.25)},yscale=-0.0017, xscale=0.0017, inner sep=0pt, outer sep=0pt]

\path[fill=ce73900,even odd rule] (27.5080,312.9100) .. controls (-40.4840,208.5300) and (41.6130,199.2100) .. (20.6720,56.3000) .. controls (72.0330,123.7200) and (86.3630,206.4200) .. (107.9330,230.8700) .. controls (122.8130,250.0800) and (130.1430,273.2700) .. (128.1430,297.0900) .. controls (127.9430,315.9400) and (151.7730,338.7900) .. (145.2530,356.3900) .. controls (110.6130,328.6400) and (51.0240,346.7300) .. (27.5130,312.9100) -- cycle;

\path[fill=ce73900,even odd rule] (130.3300,343.3900) .. controls (130.7200,323.4000) and (106.5300,314.7300) .. (96.8000,299.1200) .. controls (80.6900,287.1900) and (75.5900,268.6300) .. (75.2300,250.3500) .. controls (73.9200,221.1200) and (99.9800,46.8100) .. (137.9400,0.0000) .. controls (123.3900,71.8800) and (159.5100,167.6400) .. (181.8200,199.3200) .. controls (196.9200,223.9300) and (202.6000,252.2700) .. (197.1600,280.1600) .. controls (194.4900,302.4000) and (176.0600,315.7700) .. (166.0900,335.7200) .. controls (156.1200,347.7400) and (142.5500,348.8800) .. (130.3300,343.3900) -- cycle;

\path[fill=ce73900,even odd rule] (144.1100,353.1700) .. controls (149.9000,335.8100) and (138.4500,308.2800) .. (134.8400,292.1700) .. controls (125.0600,277.6700) and (126.0000,260.1500) .. (130.7100,244.0900) .. controls (137.6500,218.2200) and (138.8000,96.1100) .. (250.7400,38.7700) .. controls (163.9200,123.9500) and (221.5700,193.1800) .. (230.9700,226.5200) .. controls (236.4600,251.8500) and (233.6500,278.2500) .. (221.2700,301.0400) .. controls (204.1800,332.4900) and (158.5400,349.8300) .. (144.1100,353.1700) -- cycle;

\path[fill=ce78c00,even odd rule] (125.6700,342.9900) .. controls (125.9600,326.6700) and (107.6700,319.5900) .. (100.3100,306.8500) .. controls (88.1200,297.1100) and (84.2600,281.9600) .. (83.9900,267.0400) .. controls (83.0000,243.1700) and (107.3100,116.9900) .. (126.8200,62.6500) .. controls (113.5100,185.8300) and (147.7400,199.5100) .. (164.6100,225.3700) .. controls (176.0400,245.4700) and (180.3300,268.6000) .. (176.2100,291.3700) .. controls (174.1900,309.5300) and (165.4800,326.0800) .. (157.9400,342.3700) .. controls (150.4000,352.1800) and (134.9100,347.4700) .. (125.6700,342.9900) -- cycle;

\path[fill=ce78c00,even odd rule] (55.5160,319.6100) .. controls (-14.3250,247.9000) and (61.7730,242.9100) .. (42.4030,144.4300) .. controls (65.4060,190.8500) and (84.4830,225.7200) .. (92.8530,258.6800) .. controls (105.2530,270.3100) and (148.0830,341.3600) .. (145.5130,354.2400) .. controls (118.5230,338.8700) and (75.6230,340.2500) .. (55.5160,319.6100) -- cycle;

\path[fill=ce7e100,even odd rule] (69.4630,315.2900) .. controls (27.6850,275.6300) and (58.8850,252.3400) .. (58.5110,202.8100) .. controls (70.0630,246.1600) and (106.5930,256.8400) .. (106.1230,282.8900) .. controls (112.1530,290.6800) and (143.7530,346.2300) .. (141.1130,353.3700) .. controls (127.0630,342.1200) and (85.2930,330.3200) .. (69.4630,315.2900) -- cycle;

\path[fill=ce78c00,even odd rule] (136.6100,309.9400) .. controls (129.2700,299.0600) and (129.9800,285.9200) .. (133.5200,273.8700) .. controls (138.7200,254.4600) and (161.5900,146.5500) .. (194.2600,123.0800) .. controls (170.6700,167.4400) and (201.6900,235.6700) .. (208.7400,260.6800) .. controls (212.8600,279.6900) and (214.0100,302.1700) .. (201.4600,316.6000) .. controls (182.5100,338.3800) and (158.5800,351.0700) .. (145.1700,354.1100) .. controls (136.6500,360.0900) and (139.3200,322.0300) .. (136.6100,309.9400) -- cycle;

\path[fill=ce7e100,even odd rule] (124.8100,344.1900) .. controls (125.4000,334.2000) and (114.3800,329.4100) .. (110.1900,321.4200) .. controls (102.9700,315.1600) and (100.9900,305.7800) .. (101.1900,296.6400) .. controls (101.1800,282.0000) and (100.6800,190.7600) .. (119.2200,168.0800) .. controls (111.0200,203.7400) and (141.9100,256.8900) .. (151.6000,273.1500) .. controls (158.0900,285.7400) and (160.1400,300.0100) .. (157.0500,313.8400) .. controls (155.3600,324.9100) and (149.6100,334.8300) .. (144.5900,344.6100) .. controls (139.7200,350.4300) and (130.3600,347.1600) .. (124.8100,344.1900) -- cycle;

\path[fill=ce7e100,even odd rule] (133.6300,345.4600) .. controls (138.3800,338.5100) and (132.4900,330.2800) .. (132.9300,322.7100) .. controls (130.4300,315.0600) and (133.0500,307.4400) .. (137.1600,300.9300) .. controls (143.4800,290.3600) and (154.6000,211.8300) .. (177.7900,203.4800) .. controls (156.4400,225.6600) and (193.5100,284.9700) .. (193.4700,300.9000) .. controls (192.7100,312.7900) and (178.2400,328.8600) .. (170.0300,337.5100) .. controls (164.0200,344.7700) and (152.1200,351.7400) .. (144.2600,356.6300) .. controls (138.2300,352.9700) and (136.3500,350.0100) .. (133.6300,345.4600) -- cycle;

\path[fill=cffff9e,even odd rule] (125.6300,337.5100) .. controls (110.8700,314.2700) and (116.3500,288.3400) .. (122.6200,260.0600) .. controls (125.4800,279.2500) and (137.3900,302.5400) .. (144.1200,310.1800) .. controls (148.7700,316.1700) and (151.0600,323.4100) .. (150.4300,330.8500) .. controls (150.3700,336.7400) and (147.3800,343.8700) .. (145.3500,349.3600) .. controls (134.5300,340.7000) and (132.9700,348.0700) .. (125.6300,337.5100) -- cycle;

\path[fill=cffff9e,even odd rule] (142.2200,325.9900) .. controls (153.6500,300.9400) and (142.3300,281.2300) .. (168.5300,261.5800) .. controls (160.7100,279.7900) and (178.6000,308.5500) .. (176.6500,318.5300) .. controls (173.3800,335.3000) and (150.7700,347.7800) .. (145.0500,349.0700) .. controls (146.4200,335.2800) and (137.3700,337.9000) .. (142.2200,325.9900) -- cycle;

\path[fill=cffff9e,even odd rule] (60.6030,244.0200) .. controls (122.0500,321.7300) and (120.6900,282.5000) .. (145.0600,349.0500) .. controls (28.9100,297.3400) and (81.0430,302.0000) .. (60.6030,244.0200) -- cycle;

\end{scope}}},
fire/.pic={code={\input{fire.tikz}}},
treasure/.pic={code={\input{treasure.tikz}}},
goal/.pic={code={\definecolor{c030104}{RGB}{3,1,4}

\begin{scope}[shift={(-0.28,0.28)},yscale=-0.028, xscale=0.028, inner sep=0pt, outer sep=0pt]
  \path[fill=c030104] (10.0990,3.9960) -- (10.9970,5.4490) -- (9.3720,6.6400) -- (8.4470,5.1440) -- (7.1410,5.9520) -- (6.3320,4.6450) -- (7.6400,3.8360) -- (8.4490,5.1420) -- (10.0990,3.9960) -- cycle(10.1810,7.9460) -- (9.3740,6.6400) -- (8.0660,7.4480) -- (8.8730,8.7540) -- (10.1810,7.9460) -- cycle(17.6670,5.1150) -- (17.5230,5.4270) .. controls (17.2630,6.8570) and (16.3770,7.8150) .. (15.7180,8.2220) .. controls (14.2920,9.1030) and (12.8150,8.5750) .. (12.7540,8.5530) .. controls (11.7960,8.3500) and (10.9400,8.4660) .. (10.2340,8.9030) .. controls (8.9260,9.7120) and (8.5530,11.3700) .. (8.5500,11.3890) -- (8.4630,11.7960) -- (13.6450,20.1740) -- (12.8060,20.6950) -- (3.5770,5.7750) .. controls (3.4000,5.8450) and (3.1930,5.7870) .. (3.0890,5.6190) .. controls (2.9710,5.4280) and (3.0300,5.1800) .. (3.2200,5.0620) -- (3.9140,4.6340) .. controls (4.1020,4.5160) and (4.3520,4.5750) .. (4.4700,4.7660) .. controls (4.5680,4.9250) and (4.5370,5.1200) .. (4.4140,5.2510) -- (4.6630,5.6550) -- (4.6690,5.6420) .. controls (5.2110,4.3270) and (5.9010,3.4040) .. (6.7190,2.9000) .. controls (8.2030,1.9830) and (9.5590,2.7460) .. (9.6160,2.7790) .. controls (10.7760,3.4550) and (11.6020,3.2370) .. (12.0900,2.9350) .. controls (12.2990,2.8050) and (12.4090,2.6790) .. (12.4090,2.6790) .. controls (13.5550,1.6740) and (14.0160,0.5360) .. (14.2190,0.1200) .. controls (14.4210,-0.2950) and (14.7820,0.5030) .. (14.7820,0.5030) -- (17.6670,5.1150) -- cycle(14.0490,1.2090) -- (14.2580,1.5470) -- (14.7710,1.1470) -- (14.3610,0.4850) .. controls (14.2770,0.6520) and (14.2010,0.9560) .. (14.0490,1.2090) -- cycle(8.4420,10.6000) -- (8.2070,10.7460) -- (8.3180,10.9230) .. controls (8.3500,10.8300) and (8.3930,10.7170) .. (8.4420,10.6000) -- cycle(12.6910,8.1880) -- (12.5110,7.8970) -- (12.3350,8.1350) .. controls (12.4520,8.1480) and (12.5720,8.1640) .. (12.6910,8.1880) -- cycle(13.9580,7.3120) -- (13.2510,6.1700) -- (14.6840,5.1600) -- (13.8750,3.8540) -- (12.4440,4.8660) -- (11.6050,3.5110) .. controls (11.4370,3.5510) and (11.2490,3.5750) .. (11.0420,3.5730) -- (10.1000,3.9970) -- (9.5820,3.1630) .. controls (9.5340,3.1360) and (9.4870,3.1120) .. (9.4380,3.0840) .. controls (9.4220,3.0750) and (9.2970,3.0050) .. (9.0920,2.9370) -- (7.6390,3.8360) -- (7.1580,3.0610) .. controls (7.0750,3.1000) and (6.9900,3.1470) .. (6.9040,3.2000) .. controls (6.5690,3.4070) and (6.2590,3.6970) .. (5.9700,4.0600) -- (6.3310,4.6460) -- (5.1840,5.3540) .. controls (5.1320,5.4650) and (5.0810,5.5750) .. (5.0310,5.6930) -- (5.7380,6.8190) -- (7.1400,5.9520) -- (8.0660,7.4470) -- (6.6750,8.3080) -- (7.4920,9.6080) -- (8.8730,8.7540) -- (9.2240,9.3190) .. controls (9.4510,9.0550) and (9.7210,8.8050) .. (10.0490,8.6010) .. controls (10.1810,8.5210) and (10.3180,8.4540) .. (10.4570,8.3910) -- (10.1810,7.9450) -- (11.8130,6.7690) -- (10.9970,5.4490) -- (12.4420,4.8660) -- (13.2490,6.1710) -- (11.8140,6.7690) -- (12.4830,7.8500) -- (13.9580,7.3120) -- cycle(17.1470,4.9510) -- (17.1260,4.9170) -- (16.6240,5.3730) -- (15.9260,4.2470) -- (16.4520,3.8380) -- (15.5810,2.4440) -- (15.1170,2.9400) -- (14.2590,1.5500) -- (13.5020,2.0180) .. controls (13.3690,2.1940) and (13.2210,2.3670) .. (13.0640,2.5370) -- (13.8760,3.8520) -- (15.1190,2.9390) -- (15.9260,4.2470) -- (14.6850,5.1590) -- (15.3920,6.3010) -- (13.9960,7.3730) -- (14.5720,8.3030) .. controls (14.8840,8.2380) and (15.2110,8.1210) .. (15.5330,7.9220) .. controls (15.7380,7.7950) and (15.9300,7.6400) .. (16.1090,7.4610) -- (15.3920,6.3020) -- (16.6240,5.3730) -- (17.0340,6.0360) -- (17.1470,4.9510) -- cycle;

\end{scope}}},
bot/.pic={code={\definecolor{botcolor}{RGB}{61,61,160}

\begin{scope}[shift={(-0.28,0.25)},yscale=-0.018, xscale=0.018, inner sep=0pt, outer sep=0pt]
\path[fill=botcolor] (1.9990,9.9980) .. controls (0.8950,9.9980) and (0.0000,10.7800) .. (0.0000,11.7470) -- (0.0000,19.2280) .. controls (0.0000,20.1940) and (0.8950,20.9780) .. (1.9990,20.9780) .. controls (3.1020,20.9780) and (3.9980,20.1940) .. (3.9980,19.2280) -- (3.9980,11.7470) .. controls (3.9980,10.7810) and (3.1020,9.9980) .. (1.9990,9.9980) -- cycle(17.1190,2.6820) -- (18.8660,0.9400) .. controls (19.0820,0.7250) and (19.0820,0.3780) .. (18.8660,0.1620) .. controls (18.6500,-0.0530) and (18.3020,-0.0530) .. (18.0860,0.1620) -- (15.9360,2.3080) .. controls (15.1670,2.1200) and (14.3560,2.0160) .. (13.5120,2.0160) .. controls (12.6780,2.0160) and (11.8740,2.1190) .. (11.1140,2.3020) -- (8.9680,0.1620) .. controls (8.7520,-0.0530) and (8.4030,-0.0530) .. (8.1880,0.1620) .. controls (7.9720,0.3770) and (7.9720,0.7250) .. (8.1880,0.9400) -- (9.9260,2.6730) .. controls (7.0260,3.7850) and (5.0110,6.1960) .. (5.0110,8.9980) .. controls (5.0110,9.0320) and (22.0110,9.0000) .. (22.0110,8.9980) .. controls (22.0120,6.2030) and (20.0080,3.7980) .. (17.1190,2.6820) -- cycle(11.0110,7.0030) .. controls (10.4580,7.0030) and (10.0110,6.5570) .. (10.0110,6.0050) .. controls (10.0110,5.4540) and (10.4580,5.0070) .. (11.0110,5.0070) .. controls (11.5640,5.0070) and (12.0110,5.4540) .. (12.0110,6.0050) .. controls (12.0120,6.5560) and (11.5620,7.0030) .. (11.0110,7.0030) -- cycle(16.0120,7.0030) .. controls (15.4590,7.0030) and (15.0120,6.5570) .. (15.0120,6.0050) .. controls (15.0120,5.4540) and (15.4590,5.0070) .. (16.0120,5.0070) .. controls (16.5650,5.0070) and (17.0120,5.4540) .. (17.0120,6.0050) .. controls (17.0120,6.5560) and (16.5640,7.0030) .. (16.0120,7.0030) -- cycle(25.0020,9.9980) .. controls (23.8980,9.9980) and (23.0020,10.7800) .. (23.0020,11.7470) -- (23.0020,19.2280) .. controls (23.0020,20.1940) and (23.8980,20.9780) .. (25.0020,20.9780) .. controls (26.1060,20.9780) and (27.0000,20.1930) .. (27.0000,19.2280) -- (27.0000,11.7470) .. controls (27.0000,10.7810) and (26.1040,9.9980) .. (25.0020,9.9980) -- cycle(5.0110,21.9670) .. controls (5.0110,23.6170) and (6.3490,24.9530) .. (8.0010,24.9590) -- (8.0010,30.2010) .. controls (8.0010,31.1680) and (8.8960,31.9510) .. (10.0000,31.9510) .. controls (11.1030,31.9510) and (11.9980,31.1680) .. (11.9980,30.2010) -- (11.9980,24.9590) -- (15.0020,24.9590) -- (15.0020,30.2010) .. controls (15.0020,31.1680) and (15.8970,31.9510) .. (17.0010,31.9510) .. controls (18.1040,31.9510) and (18.9990,31.1680) .. (18.9990,30.2010) -- (18.9990,24.9590) -- (19.0120,24.9590) .. controls (20.6690,24.9590) and (22.0120,23.6190) .. (22.0120,21.9670) -- (22.0120,9.9950) -- (5.0120,9.9950) -- (5.0120,21.9670) -- cycle;

\end{scope}}}
}
The labeling function for this particular environment maps grid locations $(1, 0)$ and $(2, 0)$ to ``fire'' (\tikz[scale=0.5, transform shape, baseline={-0.45em}]{\pic {lava};}) and $(3, 0)$ to ``goal'' (\tikz[scale=0.5, transform shape, baseline={-0.35em}]{\pic {goal};}).
Each transition of the simple reward machine is labeled by a tuple of truth values of the two propositions (``fire'' and ``goal'').  
For example, $u_1$ transits to $u_2$ and produces a reward of $1$ if the environment's current state is labeled as ``goal'' but not ``fire''.

\paragraph{The Simple Reward Machine Objective}
Formally, a simple reward machine $R$ specifies an environment\Hyphdash generic objective $\functiontype[\denotation{R}]{\infseq{\left(2^\Pi\right)}}{\reals}$ given by:
\begingroup\small\begin{equation*}
\denotation{R}(w) = \sum_{k=1}^\infty \gamma^k \delta_r(u_k, u_{k+1}) ,
\quad \forall k \ge 0.\, u_{k+1} = \delta_u(u_k, \windex{w}{k}).
\end{equation*}\endgroup
The set of features $F$ corresponding to this environment-generic objective is the possible truth values of the propositions $\Pi$, that is $F = 2^\Pi$. 
The labeling function classifies each environment's current state and action to these features.

\section{Condition for PAC-Learnability}
\label{sec:uniform_continuity_implies_pac}

This section presents our main result: sufficient conditions for an objective's learnability. 
The first two subsections analyze learnability in the information-theoretic setting.
Specifically, we show that an objective given as an oracle is PAC-learnable if it is uniformly continuous. 
The next two subsections analyze learnability in the computation-theoretic setting.
Specifically, using a standard result from computational analysis, we show that a computable objective is PAC-learnable.
\Cref{sec:proof-of-unnecessity} complements our result by showing that our conditions are sufficient but not necessary.

\subsection{Uniform Continuity}
We first recall the following standard definition of a uniformly continuous function.

\begin{definition}[Uniformly Continuous Function]
A function $\functiontype[f]{X}{Y}$ with metric spaces $(X, d_X)$ and $(Y, d_Y)$ is uniformly continuous if, for any $\epsilon > 0$, there exists $\delta > 0$ so that $f$ maps $\delta$-close elements in the domain to $\epsilon$-close elements in the image \footnote{Note that textbook definitions commonly use ${<}$ instead of ${\le}$:
our definition is equivalent. We use $\le$ to match with the comparison operators in the PAC definitions.}
:
\begingroup\small\begin{equation*}
\begin{split}
\forall \epsilon > 0. \exists \delta > 0. \forall x_1 \in X. \forall x_2 \in X:\\
d_X(x_1, x_2) \le \delta \implies d_Y(f(x_1), f(x_2)) \le \epsilon. 
\end{split}
\end{equation*}
\endgroup
\end{definition}

To specialize the above definition to an objective, we next note the metric space of the domain of an objective.
An objective's domain is the set of infinite-length sequences $\infseq{X}$, where $X = (S \times A)$ for an environment-specific objective and $X = F$ for an environment-generic objective. 
The domain forms a metric space by the standard distance function $d_{X^\omega}(w_1, w_2) = 2^{-L_\text{prefix}(w_1, w_2)}$, where $L_\text{prefix}(w_1, w_2)$ is the length of the longest common prefix of $w_1$ and $w_2$ \citep{tlhierarchy}.
We are now ready to specialize the definition of uniform continuity to objectives.
\begin{definition}[Uniformly Continuous Objective]
An objective (environment\Hyphdash specific or environment\Hyphdash generic) $\functiontype[f]{\infseq{X}}{\reals}$ is uniformly continuous if, for any $\epsilon\,{>}\,0$, there exists a finite horizon $H$ so that the objective maps all infinite-length sequences sharing the same prefix of length $H$ to $\epsilon$-close values:
\begingroup\small\begin{equation*}
\begin{split}
\forall \epsilon > 0. \exists H \in \naturals. \forall w \in X^\omega. \forall w' \in X^\omega: \\
L_\text{prefix}(w, w') \ge H \implies |f(w) -  f(w')| \le \epsilon.
\end{split}
\end{equation*}
\endgroup
\end{definition}

Note that since the domain of an objective is compact,  the Heine–Cantor theorem guarantees that a continuous objective is also uniformly continuous.
This paper only uses uniform continuity since it is more relevant to our theorem and proof. Nonetheless, theorems presented in the following section also hold for continuous objectives.

\subsection{Continuity Implies PAC-learnability}
\label{sec:continuity-implies-pac-learnability}

\paragraph{Environment-specific Objectives.}
We give a sufficient condition for a learnable environment-specific objective:
\begin{theorem}
\label{thm:environment-specific-pac-learnable-iff-uniform-continuous}
An environment-specific objective $\kappa$ is $\kappa$-PAC-learnable in the information-theoretic setting if it is uniformly continuous.
\end{theorem}

We will prove the theorem by constructing a $\kappa$-PAC \reinforcementlearning algorithm for any uniformly continuous $\kappa$.
To that end, we reduce $\kappa$ to a finite-horizon cumulative rewards problem; 
we then prove the theorem by invoking an existing PAC \reinforcementlearning algorithm for finite-horizon cumulative rewards problems.

\begin{proof}[Proof of \Cref{thm:environment-specific-pac-learnable-iff-uniform-continuous}]

For any $\epsilon' > 0$, since $\kappa$ is uniformly continuous, there exists a bound $H$ such that infinite-length sequences sharing a length-$H$ prefix are all mapped to $\epsilon'$-close values. 

For concreteness, let us pick any $\dot{s}\,{\in}\,S$ and any $\dot{a}\,{\in}\,A$. 
For each length-$H$ sequence $u\,{\in}\,(S \times A)^H$, we pick the representative infinite-length sequence $\wrational{u}{(\dot{s}, \dot{a})}$ that starts with the prefix $u$ and ends in an infinite repetition of $(\dot{s}, \dot{a})$.
Using these representatives, we construct a finite-horizon rewards objective $\tilde{\kappa}_{\epsilon'}$ of horizon $H$. 
The construction assigns each infinite-length sequence with the value of the original $\kappa$ at the chosen representative.
That is, let $\wprefix{w}{H}$ denote the length-$H$ prefix of $w$, we define $\tilde{\kappa}_{\epsilon'}$ as:
\begingroup\small\begin{equation*}
\tilde{\kappa}_{\epsilon'}(w) \triangleq \kappa(\wrational{\wprefix{w}{H}}{(\dot{s}, \dot{a})}),\,\, \forall w \in \infseq{(S \times A)}.
\end{equation*}\endgroup

By construction, $\tilde{\kappa}_{\epsilon'}$ is $\epsilon'$-close to $\kappa$, meaning that for any infinite-length input, their evaluations differ by at most $\epsilon'$:
\begingroup\small\begin{equation*}
|\tilde{\kappa}_{\epsilon'}(w) - \kappa(w)| \le \epsilon', \,\, \forall w \in \infseq{(S\times A)}.
\end{equation*}\endgroup
Thus, an $\epsilon'$-optimal policy for $\tilde{\kappa}_{\epsilon'}$ is $2\epsilon'$-optimal for $\kappa$. 

We then reduce the approximated objective $\tilde{\kappa}_\epsilon$, which assigns a history-dependent reward at the horizon $H$, into a finite-horizon cumulative rewards objective, which assigns a history-independent reward at each step.
To that end, we lift the state space to \begingroup\small$U\,{=}\, \bigcup_{t=1}^{H} (S\times A)^t$\endgroup.
Each state \begingroup\small$u_t \,{=}\,(S\times A)^t$\endgroup at step $t$ in the lifted state space is the length-$t$ history of states and actions encountered in the environment. 
For any state before step $H$, we assign a reward of zero.
For any state $u_H \,{=}\, (S\times A)^H$ at step $H$, we assign a reward of \begingroup\small$\tilde{\kappa}_\epsilon(\wrational{u_H}{(\dot{s}, \dot{a})})$\endgroup.
The lifted state space and the history-independent reward function above form the desired finite-horizon cumulative rewards problem.

\citet{ipoc19} introduced ORLC, a PAC \reinforcementlearning algorithm for finite-horizon cumulative rewards problems.%
\footnote{ORLC provides an individual policy certificates (IPOC) guarantee. \citeauthor{ipoc19} showed that IPOC implies our PAC definition, which they called ``supervised-learning style PAC''.}
Applying ORLC to the above finite-horizon cumulative rewards problem produces an $2\epsilon'$-optimal policy for $\kappa$.
Finally, for any $\epsilon$, choosing $\epsilon' \;{=}\; \frac{\epsilon}{2}$ gives a $\kappa$-PAC \reinforcementlearning algorithm for $\kappa$. 
\end{proof}

\paragraph{Environment-generic Objectives.}
\Cref{thm:environment-specific-pac-learnable-iff-uniform-continuous} states a sufficient condition for when an environment-specific objective is PAC-learnable. 
The following corollary generalizes the condition to environment-generic objectives.
\begin{corollary}\label{thm:environment-generic-pac-learnable-if-uniform-continuous}
An environment-generic objective $\xi$ is $\xi$-PAC-learnable in the information-theoretic setting if $\xi$ is uniformly continuous.
\end{corollary}
To the end of proving \Cref{thm:environment-generic-pac-learnable-if-uniform-continuous}, we first observe the following lemma, which we prove in \Cref{sec:label_preserves_uniform_continuity_proof}.
\begin{lemma}\label{lemma:label_preserves_uniform_continuity}
If an environment-generic objective is uniformly continuous, then, for all labeling functions, the induced environment\Hyphdash specific objective is also uniformly continuous.
\end{lemma}

With \Cref{lemma:label_preserves_uniform_continuity}, \Cref{thm:environment-generic-pac-learnable-if-uniform-continuous} is straightforward.
Since each induced environment-specific objective $\kappa$ is uniformly continuous, each $\kappa$ is $\kappa$-PAC-learnable by \Cref{thm:environment-specific-pac-learnable-iff-uniform-continuous}.
Thus, the objective $\xi$ is $\xi$-PAC-learnable by definition.

\subsection{Computability}
\label{sec:computability}

We now define the computability of an objective $\functiontype[f]{\infseq{X}}{\reals}$.
The standard definition of computability of such functions depends on Type-2 Turing machines~\citep[Chapter~2, Definition 2]{weihrauch2000computable} and a representation of the reals by an infinite sequence of rational approximations, called the Cauchy-representation \citep[Chapter~3, Definition 3]{weihrauch2000computable}.
Informally, a Type-2 Turing machine is a Turing machine with an infinite-length input tape and a one-way infinite-length output tape. The machine reads the input tape and computes forever writing to the output tape. 
\begin{definition}[Computable Objective]
\label{def:computable-objective}
An objective $f$ is computable if a Type-2 Turing machine reads $w$ from the input tape and writes to the output tape a fast-converging Cauchy sequence $[q_0, q_1, \dots] \in \infseq{\rationals}$ of rational approximations to $f(w)$, that is:
\begingroup\small$\forall n \in \naturals$, $|f(w) - q_{n}| \le 2^{-n}$.\endgroup
\end{definition}
When proving computability, this definition is tedious to work with since it requires implementing the function on a Turing machine.
Instead, we will use pseudocode to formulate an algorithm that takes in an infinite-stream input $w$ and a natural number $n$ and outputs the $n$-th rational approximation $q_n$. 
Repeatedly invoking the algorithm by enumerating $n$ produces the Cauchy sequence of rational approximations.

A classic result in computable analysis is that computable functions are continuous \citep[Theorem~2.5 and 4.3]{weihrauch2000computable}. 
Since an objective's domain is compact, by the Heine-Cantor theorem, this result also holds for uniform continuity. 
Even stronger, the following theorem, modified from \citet[Theorem~6.4]{weihrauch2000computable} for our context, guarantees that for a computable objective, for any rational $\epsilon > 0$, we can compute a horizon $H$ that satisfies the definition of uniform continuity.
Define {\em the modulus of continuity} of an objective as a function $\functiontype[m]{\rationals}{\naturals}$ that satisfies 
\begingroup\small$\forall \epsilon \in \rationals$, $\forall w_1 \in \infseq{X}, \forall w_2 \in \infseq{X}: L_\text{prefix}\left(w_1, w_2\right) \ge m(\epsilon) \implies \left|f\left(w_1\right) - f\left(w_2\right)\right| \le \epsilon$\endgroup.
Then, we have:
\begin{theorem}
\label{thm:computable-implies-uc}
A computable objective is uniformly continuous. Further, its modulus of continuity $m$ is computable.
\end{theorem}
For completeness, \Cref{sec:computing-modulus-of-continuity} gives pseudocode that computes the modulus of continuity for any computable objective specified by the interface described above.

\subsection{Computability Implies PAC-learnability}

We now extend our result in \Cref{sec:continuity-implies-pac-learnability} from the information-theoretic to the computation-theoretic setting. 
\begin{theorem}
\label{thm:objective-pac-learnable-if-computable}
An (environment-generic or environment\Hyphdash specific) objective $f$ is $f$-PAC-learnable in the computation-theoretic setting if $f$ is computable.
\end{theorem}
\begin{proof}
Combining theorems in \Cref{sec:continuity-implies-pac-learnability} and \Cref{thm:computable-implies-uc}, a computable objective $f$ is uniformly continuous, therefore $f$-PAC-learnable in the information-theoretic setting.
In the computational-theoretic setting, we need to further construct a computable \reinforcementlearning algorithm.
Note that our proof of \Cref{thm:environment-generic-pac-learnable-if-uniform-continuous} is already constructive of an algorithm. However, we need to:  
\begin{itemize}[left=0pt]
\item compute the bound $H$ from the given $\epsilon'$ and
\item computably evaluate the approximated objective $\tilde{\kappa}_{\epsilon'}$.
\end{itemize}
A computable objective resolves both points: 
\begin{itemize}[left=0pt]
\item By \Cref{thm:computable-implies-uc}, the bound $H$ is computable for any $\epsilon$. 
\item Evaluating the approximate objective is computable, since the approximated objective only depends on the length-$H$ prefix of the input.\qedhere
\end{itemize}
\end{proof}
\Cref{sec:concrete-pac-rl-algorithm} provides pseudocode for an $f$-PAC \reinforcementlearning for any computable objective $f$.

\section{Theorem Applications}
\label{sec:theorem-apps}

This section applies the core theorem and corollary to two objectives in the existing literature and proves each objective's PAC-learnability.
Due to space, we give the third objective from \citet{gltl17} and prove its PAC-learnability in \Cref{sec:gltl}.

\subsection{Reward Machine}

\paragraph{Proof of PAC-learnability} 
We prove that the reward-machine objective reviewed in \Cref{sec:reward_machine_example} is PAC-learnable.
\begin{proposition}
The objective $\denotation{R}$ of a simple reward machine $R$ is $\denotation{R}$-PAC-learnable.
\end{proposition}
\begin{proof}
By \Cref{thm:objective-pac-learnable-if-computable}, it is sufficient to show that a simple reward-machine objective is computable. 
Consider the pseudocode with Python-like syntax in \Cref{lst:simple-reward-machine-objective-impl}. 
\begin{pseudocode}[caption={Computation of the simple reward objective},label={lst:simple-reward-machine-objective-impl}]
# Given reward machine (*$(U, \delta_u, \delta_r, u_0, \gamma)$*)
def SRM(w: (*$\infseq{(2^\Pi)}$*), n: (*$\naturals$*)) -> (*$\rationals$*):
  u: (*$U$*), value: (*$\rationals$*)  = (*$u_0$*), 0
  rmax: (*$\rationals$*) = max(abs((*$\delta_r(\text{u1}, \text{u2})$*)) for u1, u2 in (*$U^2$*)))
  H: (*$\naturals$*) = (log2floor(1 - (*$\gamma$*)) -n - log2ceil(rmax))) \
            / log2ceil((*$\gamma$*))
  for k in range(H):
    u' = (*$\delta_u$*)(u, w[k])
    value += (*$\gamma$*)**k * (*$\delta_r$*)(u, u')
    u = u'
  return value
\end{pseudocode}
\vspace{1em}

\Cref{lst:simple-reward-machine-objective-impl} defines an algorithm for computing the simple reward-machine objective.
It first initializes the state variable \texttt{u} to the initial state $u_0$.
It then computes a horizon $H = \left(\lfloor\log_2 (1 - \gamma) \rfloor - n - \lceil\log_2 r_{\max}\rceil\right) / \lceil\log_2\gamma\rceil$, where \begingroup\small$r_{\max} = \max\left(\left|\delta_r(\cdot)\right|\right)$ \endgroup is the maximum magnitude of all possible rewards.
It iterates through the first $H$ indices of the input and transits the reward machine's state according to the transitions $\delta_u$. 
For each input $w$ and $n$, the algorithm accumulates the discounted cumulative rewards truncated to the first $H$-terms: $\sum_{k=0}^{H-1} \gamma^k \delta_r(u_k, u_{k+1})$.

By definition of a computable objective, we need to show that the returned values for all $n$ form a fast-converging Cauchy sequence: $\forall n \in \naturals, |\texttt{SRM}(w, n) - \denotation{R}(w)| \le 2^{-n}$.
To see this, let
$\Delta \triangleq \left|\texttt{SRM}(w, n) - \denotation{R}\left(w\right)\right| = 
\left|\sum_{k=H}^{\infty} \gamma^k \delta_r(u_k, u_{k+1})\right|
$.
Then, we have $\Delta \le \nicefrac{r_{\max} \cdot \gamma^H}{1 - \gamma}$ by upper bounding the rewards by $r_{\max}$, then simplifying the infinite sum into a closed form.
By plugging in the value of $H$ and simplifying the inequality, we have $\Delta \le 2^{-n}$.
Thus, the objective is computable and $\denotation{R}$-PAC-learnable.\qedhere

\end{proof}

\subsection{LTL-in-the-limit Objectives}

Linear temporal logic (LTL) objectives are measurable Boolean objectives that live in the first two-and-half levels of the Borel hierarchy \citep{tlhierarchy}.
Various works \citep{hasanbeig2019reinforcement,omegaregularrl19,bozkurt2020control} considered LTL objectives for reinforcement learning and empirical algorithms for learning. 
A common pattern of these algorithms is that they convert a given LTL formula to an intermediate specification that takes in additional hyper-parameters. 
They show that in an unreachable limit of these hyper-parameters, the optimal policy for this intermediate specification becomes the optimal policy for the given LTL formula.
We call such intermediate specifications {\em LTL-in-the-limit specifications}.
Due to space, we will focus on  \citet{bozkurt2020control} and give the objective specified by their LTL-in-the-limit-specification.
We will show that this objective is PAC-learnable.
The same process, namely writing down the LTL-in-the-limit specification and then proving that the specified objective is PAC-learnable, also applies to the approaches in \citet{dorsa14,hasanbeig2019reinforcement,omegaregularrl19}.

\paragraph{\citeauthor{bozkurt2020control}'s LTL-in-the-limit Specification}
\label{sec:ltl-in-the-limit-spec-intro}

Given an LTL formula, \citeauthor{bozkurt2020control} first convert the formula into a {\em limit deterministic Buchi Automaton (LDBA)} by a standard conversion algorithm \citep{Sickert2016LimitDeterministicBA} with two additional discount factor parameters.
An LDBA is a non-deterministic finite automaton. 
It is bipartite by two sets of states, those in an {\em initial component} and those in an {\em accepting component}.
Transitions in the automaton can only go from the initial component to the accepting component, but not the reverse. 
An LDBA is ``deterministic in the limit'': it only has non-deterministic $\epsilon$-transitions in the initial component, but it is deterministic in the accepting component. The formal definition of LDBA is:

\begin{definition}[LDBA]
For an LTL formula over propositions $\Pi$, an LDBA converted from the formula is a tuple $(U, \mathcal{E}, \delta_u, u_0, B)$, where $U$ is a finite set of states, $\functiontype[\delta_u]{\left(U \times (2^\Pi \cup \{\epsilon\})\right)}{2^U}$ is a non-deterministic transition function, $u_0$ is an initial state, and $B \subseteq U$ is a set of accepting states.  
Additionally, $U$ has a bi-partition of an initial component with states $U_I$ and an accepting component with states $U_B$. An LDBA satisfies the conditions:
\begin{enumerate*}[(1)]
\item $\delta_u(u, \epsilon) = \emptyset$ for all $u \in U_B$, 
\item $\delta_u(u, 2^\Pi) \subseteq U_B$ for all $u \in U_B$, and
\item $B \subseteq U_B$. 
\end{enumerate*}
\end{definition}

The agent and environment models are similar to a simple reward machine: At each step, the agent chooses an environment's action and steps in the environment. 
A labeling function classifies the current state of the environment to a tuple of truth values of the set of propositions $\Pi$. 

At each step, an LDBA takes either a non\Hyphdash deterministic $\epsilon$-transitions (if such transition is available) or the transition along the tuple of the truth values of the propositions. 
Each time the LDBA enters an accepting state, the agent receives a reward of $1 - \gamma_1$, and discounts all future rewards by $\gamma_1$.
Each time the LDBA enters a non-accepting state, the agent receives no reward and discounts all future rewards by $\gamma_2$.
An oracle controls the $\epsilon$-transitions.
In words, the objective is to maximize the (state-dependent) discounted cumulative rewards, assuming the oracle always makes the optimal choice that helps to maximize the cumulative rewards.

\paragraph{\citeauthor{bozkurt2020control}'s LTL-in-the-limit Objective}
\citeauthor{bozkurt2020control}'s LTL-in-the-limit specification is a tuple $(L, \gamma_1, \gamma_2)$: the LDBA $L$ and the two hyper-parameters $\gamma_1, \gamma_2 \in \rationals$.
It specifies an environment-generic objective $\functiontype[\denotation{(L, \gamma_1, \gamma_2)}]{\infseq{(2^\Pi)}}{\reals}$.
Let $\mathcal{E}^+ = \mathcal{E} \cup \{\bot\}$, where $\mathcal{E}$ is the set of $\epsilon$-transitions and $\bot$ is a non-$\epsilon$-transition (i.e., following a transition with a tuple classified by the labeling function),
the objective is:
\begingroup\smaller\begin{equation}
\label{eq:bozkurt_objective_def}
\begin{split}
& \denotation{(L, \gamma_1, \gamma_2)}(w) = \max_{w_\mathcal{E} \in \infseq{\mathcal{E}}} g(w_\mathcal{E}, w) \quad \text{where } \\
& g(w_\mathcal{E}, w) = \sum_{i=1}^{\infty} R(u_i) \prod_{j=1}^{i-1} \Gamma(u_j), \\ 
& R(u) =  (1 - \gamma_1) \indicator{u \in B}, \\ 
& \Gamma(u) = \gamma_1 \indicator{u \in B} + \gamma_2 \indicator{u \not \in B} , \\
& \forall k \ge 0 \colon u_{k+1} = \delta_u(u_k, w^+_k), \\ 
& \quad t_k = \sum_{i=1}^{k} \indicator{ \windex{w_\mathcal{E}}{i} = \bot \text{ or } (u_k, \windex{w_\mathcal{E}}{i}) \not\in \delta_u }, \\
& \quad w^+_k = \begin{cases}
\windex{w}{t_k} & \text{if } \windex{w_\mathcal{E}}{k} = \bot \text{ or } (u_k, \windex{w_\mathcal{E}}{k}) \not\in \delta_u \\
\windex{w_\mathcal{E}}{k} & \text{otherwise}
\end{cases}
\end{split} .
\end{equation}
\endgroup
Here, $t_k$ is the step count of the environment when the LDBA takes its $k$-th step. 
Note that $t_k \le k$, since the environment does not step when LDBA takes an $\epsilon$-transition.
The value $\windex{w}{t_k}$ is the tuple of truth values of the input infinite-length sequence $w$ at $t_k$.
We define $w^+_k$ as the transition label taken by the LDBA at the $k$-th step: It is either
\begin{enumerate*}[(1)]
\item a tuple of truth values $\windex{w}{k}$, if  $\windex{w_\mathcal{E}}{k}$ is a non-$\epsilon$-transition or an $\epsilon$-transition that is not available from the current LDBA state $u_k$, or
\item the $\epsilon$-transition $\windex{w_\mathcal{E}}{k}$.
\end{enumerate*}
By its definition, $w^+_k$ is always a valid transition of the LDBA, and it always leads to a deterministic next state. 
Therefore, we write $u_{k+1} = \delta_u(u_k, w^+_k)$ to denote that the LDBA state $u_{k+1}$ follows this deterministic transition to the next state.

\paragraph{Proof of PAC-learnability}

We now prove that the objective specified by an LTL-in-the-limit specification in \citet{bozkurt2020control} is PAC-learnable.
Although this section aims to show an example, as we mentioned, the proof strategy here also applies to the approaches in \citet{dorsa14,hasanbeig2019reinforcement,omegaregularrl19}.

\begin{proposition}
\citeauthor{bozkurt2020control}'s LTL-in-the-limit objective $\denotation{(L, \gamma_1, \gamma_2)}$ is $\denotation{(L, \gamma_1, \gamma_2)}$-PAC-learnable.
\end{proposition}
\begin{proof}
By \Cref{thm:objective-pac-learnable-if-computable}, it is sufficient to show that \citeauthor{bozkurt2020control}'s objective is computable. 
Consider the pseudocode with Python-like syntax in \Cref{lst:ldba-objective-impl}. 
\begin{pseudocode}[caption={Computation of \citeauthor{bozkurt2020control}'s objective},label={lst:ldba-objective-impl},float=tp]
# Given LDBA (*$(U, \mathcal{E}, \delta_u, u_0, B)$*) and (*$\gamma_1, \gamma_2$*)
def bozkurt_objective(w: (*$\infseq{(2^\Pi)}$*), n: (*$\naturals$*)) -> (*$\rationals$*):
  gamma_max: (*$\rationals$*)  = max((*$\gamma_1$*), (*$\gamma_1$*))
  H: (*$\naturals$*)  = (log2floor(1 - gamma_max) - n) \
            / log2ceil(gamma_max)
  v: (*$\rationals$*) = 0
  for w_e in (*$\mathcal{E}^\text{H}$*):
    v = max(v, bozkurt_helper(H, w_e, w))
  return v

def bozkurt_helper(H: (*$\naturals$*), w_e: (*$\mathcal{E}^\text{H}$*), w: (*$S^\omega$*)) -> (*$\rationals$*):
  v: (*$\rationals$*), u: (*$U$*), discount: (*$\rationals$*)  = 0, (*$u_0$*), 1
  for k in range(H):
    if u in (*$B$*):
      reward, gamma = 1, (*$\gamma_1$*)
    else:
      reward, gamma = 0, (*$\gamma_2$*)
    v += reward * discount
    discount *= gamma
    if w_e[k] == (*$\bot$*) or (u, w_e[k]) not in (*$\delta_u$*):
      w_k_plus = w[k]
    else: 
      w_k_plus = we[k]
    u = (*$\delta_u$*)(u, w_k_plus)
  return v
\end{pseudocode}

\Cref{lst:ldba-objective-impl} gives pseudocode for computing \citeauthor{bozkurt2020control}'s objective.
The pseudocode contains two procedures.
The procedure \texttt{bozkurt\_helper} computes $g$ but truncates the sum to the first $H = \left(\lfloor\log_2(1-\max(\gamma_1, \gamma_2)\rfloor - n\right) / \lceil\log_2 \max(\gamma_1, \gamma_2)\rceil$ terms. 
The procedure \texttt{bozkurt\_objective} then computes the $n$-th rational approximation of the objective's value. 
It invokes the helper function for all $\hat{w_\epsilon} \in \mathcal{E}^n$ and calculates the value of $\max_{\hat{w_\epsilon} \in \mathcal{E}^n} \texttt{bozkurt\_helper}(\hat{w_\epsilon}, w, n)$.

\Cref{sec:bozkurt_objective_pac_proof} proves that the return values of \texttt{bozkurt\_objective} for all $n\in\naturals$ form a fast-converging Cauchy sequence: 
\begingroup\small\begin{equation*}
|\texttt{bozkurt\_objective}(w, n) - \denotation{(L, \gamma_1, \gamma_2)}(w)| \le 2^{-n}.
\end{equation*}\endgroup
Therefore, the objective is computable and consequently $\denotation{(L, \gamma_1, \gamma_2)}$-PAC-learnable.
\end{proof}

\section{Conclusion}

This work studies the PAC-learnability of general \reinforcementlearning objectives and gives the first sufficient condition of PAC-learnability of an objective. 
We use examples to show the applicability of our condition on
various existing objectives whose learnability were previously
unknown.

\paragraph{Applications to Existing Objectives}
Although we only demonstrated three examples, our theorem also applies to other objectives in the literature.
Some examples are 
\begin{enumerate*}[(1)]
\item modifications to the simple reward machine such as the (standard) reward machine \citep{rewardmachine1} (where rewards depend on not only the reward machine's state but also the environment's state) and the stochastic reward machine \citep{stochasticrewardmachine22},
\item other LTL-in-the-limit objectives \citep{dorsa14,omegaregularrl19,hasanbeig2019reinforcement}, and
\item various finite-horizon objectives \citep{smcbltl,osbert,ltlf-rl}.
\end{enumerate*}

Moreover, we gave an example objective in \Cref{sec:proof-of-unnecessity} showing our condition is sufficient but not necessary. 
However, to our knowledge, no previous objective has a similar pattern to our example. 
Therefore, we conjecture that our condition applies to most, if not all, existing PAC-learnable objectives in the literature. Nonetheless, verifying each objective's PAC-learnability is out of scope of this work.

\paragraph{Guiding The Design of New Objectives}
Our main result could also help the design of new objectives.
With our sufficient condition, researchers can create continuous and computable objectives by design, and our condition will ensure the PAC-learnability of such objectives.

\newcommand{\arxivpp}[1]{arXiv preprint: #1}

\ifseparateappendix
\else
\clearpage
\onecolumn
\numberwithin{equation}{section}
\numberwithin{definition}{section}
\numberwithin{theorem}{section}
\numberwithin{lemma}{section}
\numberwithin{proposition}{section}
\numberwithin{figure}{section}
\numberwithin{table}{section}
\numberwithin{lstlisting}{section}
\begin{appendices}
\flushcolsend
\crefalias{section}{appendix}
\crefalias{subsection}{appendix}
\section{Summary of Works on LTL-in-the-Limit Objectives}
\label{sec:summary-rl-automaton-objectives}
This section briefly reviews the literature on LTL-in-the-limit objectives. \footnote{We acknowledge the valuable input from an anonymous reviewer that helped us with this summary.}

To our knowledge, \citet{dorsa14} first proposed LTL as an objective for model-free reinforcement learning. They transformed LTL formulas into Rabin automata that give out rewards to the agent. Although the approach was appealing, \citet{omegaregularrl19} identified counterexamples demonstrating that the translation was not entirely correct. Subsequently, both \citet{omegaregularrl19} and \citet{hasanbeig2019reinforcement} proposed to use LDBA-based reward schemes, and \citet{omegaregularrl19}'s approach addressed the issues in \citet{dorsa14}.
Later, \citet{bozkurt2020control} proposed an LDBA-based reward scheme that was less sparse than previous LDBA-based reward schemes---meaning this scheme provides rewards not only at the sink states of the LDBA, but also at intermediate states. 
In the same year, \citet{omegaregularrl20} proposed a dense reward scheme and conducted experimental comparisons of various approaches.

Note that although all the above approaches attempt to use LTL as reinforcement-learning objectives, \citet{ltlnonpac22} proved that PAC learning is only possible for a subset of LTL formulas called the finitary formulas.
Nonetheless, approaches in the previous paragraph all fall into 
a common pattern that they convert a given LTL formula to an intermediate specification (Rabin automaton or LDBA) that takes in additional hyper-parameters. 
They show that in an unreachable limit of these hyper-parameters, the optimal policy for this intermediate specification becomes optimal for the given LTL formula. 
As mentioned in \Cref{sec:ltl-in-the-limit-spec-intro}, we view these approaches as introducing LTL-in-the-limit objectives and using them as proxies to the true LTL objectives.

\section{Proof of \Cref{lemma:label_preserves_uniform_continuity}}
\label{sec:label_preserves_uniform_continuity_proof}

\begin{proof}
Since $\xi$ is uniformly continuous, by definition, we have:
\begin{equation*}
\begin{split}
\forall \epsilon > 0. \exists H \in \naturals. \forall w_3 \in \infseq{F}. \forall w_4 \in \infseq{F}. \\
L_\text{prefix}(w_3, w_4) \ge H \implies |\xi(w_3) -  \xi(w_4)| \le \epsilon .
\end{split}
\end{equation*}
Let $\kappa = \mathcal{\xi} \circ \mathcal{L}$ be the environment-specific objective induced by $\xi$ and the labeling function $\mathcal{L}$.
By rewriting $w_3$ as $\mathcal{L}(w_1)$ and $w_4$ as $\mathcal{L}(w_2)$, we get:
\begin{equation}
\label{eq:proof_lemma3_eq1}
\begin{split}
\forall \epsilon > 0. \exists H \in \naturals. \forall w_1 \in \infseq{(S\times A)}. \forall w_2 \in \infseq{(S\times A)}. \\
L_\text{prefix}\left(\mathcal{L}\left(w_1\right), \mathcal{L}\left(w_2\right)\right) \ge H \implies |\kappa(w_1) -  \kappa(w_2)| \le \epsilon .
\end{split}
\end{equation}

Consider any $w \in \infseq{(S\times A)}$ and $w' \in \infseq{(S\times A)}$. 
If $w$ and $w'$ share a prefix of length $H$, 
then the labeling function also maps them to infinite-length paths sharing a prefix of length at least $H$, that is:
\begin{equation}
\label{eq:proof_lemma3_eq2}
\forall H \in \naturals. \forall w_1 \in \infseq{(S\times A)}. \forall w_2 \in \infseq{(S\times A)}. \, L_\text{prefix}(w_1, w_2) \ge H \implies L_\text{prefix}(\mathcal{L}(w_1), \mathcal{L}(w_2)) \ge H.
\end{equation}

By chaining the implications in \Cref{eq:proof_lemma3_eq1} and \Cref{eq:proof_lemma3_eq2}, we get:
\begin{equation*}
\begin{split}
\forall \epsilon > 0. \exists H \in \naturals. \forall w_1 \in \infseq{(S\times A)}. \forall w_2 \in \infseq{(S\times A)}. \\
L_\text{prefix}\left(w_1, w_2\right) \ge H \implies |\kappa(w_1) - \kappa(w_2)| \le \epsilon .
\end{split}
\end{equation*}
Therefore, $\kappa$ is uniformly continuous by definition.
\end{proof}

\section{Computing the Modulus-of-Continuity}
\label{sec:computing-modulus-of-continuity}

\Cref{lst:compute-modulus-of-continuity} gives pseudocode for computing the modulus of continuity of any computable objective given by the interface $\functiontype{\infseq{X}, \naturals}{\rationals}$ (described in \Cref{sec:computability}).

\begin{pseudocode}[label={lst:compute-modulus-of-continuity}, caption={Computation of the modulus-of-continuity of a computable objective}]
def modulus_of_continuity(objective: (*$\functiontype{\infseq{X}, \naturals}{\rationals}$*), (*$\epsilon$*): (*$\rationals$*)) -> (*$\naturals$*):
  H = 1
  while True:
    try:
      for w in (*$X^H$*):
        objective(w, log2ceil((*$\epsilon$*)))
      return H
    except OutOfBound:
      H += 1
\end{pseudocode}
The algorithm enumerates $H$ from 0. It forms finite-length words $w$ for each $X^H$ and invokes the objective on $w$ and $n = \lceil\log_2(\epsilon)\rceil$. 
If the computation of the objective attempts to read $w[k]$ for some $k$ greater than $H$, an exception is thrown. The exception terminates the enumeration of $X^H$ and returns the control to the outer loop.
The algorithm essentially finds the first $H$ such that the objective only needs to inspect the first $H$ indices of $w$ to calculate an $\epsilon$-close approximation to the objective's value.

\section{PAC Reinforcement-Learning Algorithm for Computable Objectives}
\label{sec:concrete-pac-rl-algorithm}

\Cref{lst:rl-computable-objective} gives pseudocode for a \reinforcementlearning algorithm for any computable objective given by the interface $\functiontype{\infseq{X}, \naturals}{\rationals}$ (described in \Cref{sec:computability}).
The algorithm first computes a sufficient horizon bound $H$ for achieving $\frac{\epsilon}{2}$-approximation to the objective's value. 
It then constructs the lifted MDP with finite-horizon cumulative rewards as described in the proof of \Cref{thm:environment-specific-pac-learnable-iff-uniform-continuous}. 
Finally, it invokes \texttt{rl\_finite\_horizon\_cumulative\_rewards}, an existing PAC \reinforcementlearning algorithm for finite-horizon cumulative rewards problem to obtain a $\frac{\epsilon}{2}$-optimal policy.
Overall, the obtained policy is an $\epsilon$-optimal policy to the computable objective.

\begin{pseudocode}[float={tp}, caption={Pseudocode for a \reinforcementlearning algorithm for computable objectives}, label={lst:rl-computable-objective}]
def rl_general_objective(epsilon, delta, objective, mdp, label_fn):
    epsilon' = epsilon / 2
    H = modulus_of_continuity(objective, epsilon')

    def lifted_transition(state, action):
        # current MDP state is the last state in the history
        if len(state) == 1:
          mdp_state = state[-1]
        else:
          last_action, mdp_state = state[-1]
        next_mdp_state = mdp.step(mdp_state, action) # step to sample the environment
        next_state = state + ((action, next_mdp_state), ) # append next MDP state
        return next_state

    def reward_fn(state, action):
        if len(state) == H:
            # If current state is the last state in the horizon H
            # then give reward of the value of the approximated objective
            return objective(state, epsilon')
        else:
            return 0 # reward 0 otherwise
    
    # Invoke existing PAC reinforcement-learning algorithm for finite-horizon cumulative rewards
    policy = rl_finite_horizon_cumulative_rewards(
        epsilon',
        delta
        horizon=H, 
        mdp=MDP(step=lifted_transition, 
                reward_fn=reward_fn, 
                init_state=(mdp.init_state,))
    )

    return policy
\end{pseudocode}

\section{Proof of Unnecessity}
\label{sec:proof-of-unnecessity}

We complement our result and prove that our conditions are only sufficient but not necessary. 
To that end, we give an objective that is not uniformly continuous (or computable) but is PAC-learnable. 
Consider an environment-generic objective $\functiontype[\xi]{\infseq{\{a, b\}}}{\reals}$ with features $F = \{a, b\}$, given by:
\begin{equation*}
\xi(w) = \indicator{w \neq \hat{w}} \quad \text{ where }  \hat{w} = abaabaaab\dots
\end{equation*}
That is, the objective assigns a value of $0$ for $\hat{w}$, which is an infinite-length path with a naturally increasing number of $a$s between infinitely many $b$s, and $1$ otherwise.
Any finite state DTMC has zero probability of generating $\hat{w}$, since the pattern of $\hat{w}$ necessarily requires an infinite memory to generate. 
Thus, this objective's value is $1$ for any environment and any policy. 
In other words, for any environment, all policies are equally optimal. 
Therefore, the objective is trivially PAC-learnable. 
However,  due to discontinuity at $\hat{w}$, the objective is neither uniformly continuous nor computable.

Although our condition is only sufficient, to the best of our knowledge, no existing objectives in the literature have a similar nature to the above example, which would make our conditions inapplicable.

\section{Proof of Computability of \Cref{lst:ldba-objective-impl}}
\label{sec:bozkurt_objective_pac_proof}

Let $g_{a:b}(w_\epsilon, w) \triangleq \sum_{i=a}^{b} R(u_i) \prod_{j=1}^{i-1} \Gamma(u_j) $ denote the partial sum of $g$ in \Cref{eq:bozkurt_objective_def} from $a$ to $b$.
Let $z_{h}$ be the sequence of $\epsilon$-moves of length-$h$ that maximizes $g_{0:h}$, that is: $z_{h} \triangleq \argmax_{\hat{w_\epsilon} \in \mathcal{E}^h} g_{0:h}(\hat{w_\epsilon}, w)$.
Similarly, let $z_\infty \in \infseq{\mathcal{E}}$ be the infinite-length path of $\epsilon$-moves that maximizes $g$. 
Then we can write $\Delta$ as:
\begingroup\small\begin{equation*}
\begin{split}
\Delta & \triangleq \left|\denotation{(L, \gamma_1, \gamma_2)}(w) - \texttt{bozkurt\_objective}(w, n)\right|  \\ 
&= \left|\max_{w_\epsilon \in \infseq{\mathcal{E}}} g(w_\epsilon, w) - \max_{\hat{w_\epsilon} \in \mathcal{E}^{H}} g_H(\hat{w_\epsilon}, w) \right| \\
&= \left|g(z_\infty, w) - g_{0:H}(z_H, w) \right|
\end{split}
\end{equation*}\endgroup

Observe that $g(z_\infty, w) \ge g_{0:H}(z_H, w)$.
To see this inequality, let $\tilde{w_\epsilon} \in \infseq{\mathcal{E}}$ be any infinite\Hyphdash length path of $\epsilon$-moves with $z_H$ as the prefix. Then we must have $g(\tilde{w_\epsilon}, w) \le g(z_\infty, w)$, since $z_\infty$ maximizes $g$. 
Moreover, since $g_{0:H}(z_H, w)$ is just a partial sum of $g(\tilde{w_\epsilon}, w)$ and since each term of the summation is positive (because $R$ and $\Gamma$ are postive), we have $g_{0:H}(z_H, w) \le g(\tilde{w_\epsilon}, w)$.
We chain the inequalities to get $g(z_\infty, w) \ge g_{0:H}(z_H, w)$.
Therefore we may drop the absolute value:  
$
\Delta = g(z_\infty, w) - g_{0:H}(z_H, w)
$.

We now bound $\Delta$ by bounding $g(z_\infty, w)$:
\begingroup\small\begin{equation*}
\begin{split}
\Delta &= g(z_\infty, w) - g_{0:H}(z_H, w) \\
& = g_{0:H}(z_\infty, w) + g_{H:\infty}(z_\infty, w) - g_{0:H}(z_H, w) \\
& \le g_{H:\infty}(z_\infty, w)
\end{split}
\end{equation*}
\endgroup
The second equality holds by splitting the summation in $g$. 
The last inequality holds since $z_H$ maximizes $g_{0:H}$: $g_{0:H}(z_H, w) \ge g_{0:H}(z_\infty, w)$.

Therefore, after expanding the definition of $g_{H:\infty}$, we have 
$
\Delta \le \sum_{i=H}^{\infty} R(u_i) \prod_{j=1}^{i-1} \Gamma(u_j)
$.
Since $R(u_i) < 1$ and $\Gamma(u_j) \le \max(\gamma_1, \gamma_2)$, we have $\Delta \le \sum_{i=H}^{\infty} \max(\gamma_1, \gamma_2)^{i-1}$.
Simplifying the sum, we get $\Delta \le \frac{\max(\gamma_1, \gamma_2)^{H-1}}{1 - \max(\gamma_1, \gamma_2)}$.
Finally, we plug in the value of $H = \left(\lfloor\log_2(1 - \max(\gamma_1, \gamma_2))\rfloor - n \right)/ \lceil\log_2\max(\gamma_1, \gamma_2) \rceil$ and get $\Delta \le 2^{-n}$.
Therefore, the objective is computable and consequently $\denotation{(L, \gamma_1, \gamma_2)}$-PAC-learnable.

\section{Geometric Linear Temporal Logic}
\label{sec:gltl}

\citet{gltl17} introduced geometric linear temporal logic (GLTL), a variant of linear temporal logic with expiring temporal operators.
This section formalizes the objective specified by a GLTL formula and proves that the objective is PAC-learnable.

\subsection{GLTL Specification}

A GLTL formula is built from a finite set of atomic propositions $\Pi$, logical connectives $\neg, \land, \lor$, temporal next $\ltlnext$, and expiring temporal operators  $\always_\theta$ (expiring always), $\eventually_\theta$ (expiring eventually), and $\until_\theta$ (expiring until). 
\Cref{eq:gnl_grammar} gives the grammar of a GLTL formula $\phi$ over the set of atomic propositions $\Pi$:
\begingroup
\newcommand{\alt}{\:\big|\:}%
\begin{equation}
\label{eq:gnl_grammar}
\phi \defeq 
a
\alt \neg \phi 
\alt \phi \land \phi 
\alt \phi \lor \phi
\alt \ltlnext \phi
\alt \always_\theta \phi 
\alt \eventually_\theta \phi
\alt \phi \until_\theta \phi, \; a \in \Pi, \theta \in \rationals.
\end{equation}
\endgroup
Each temporal operator (i.e., $\always$, $\eventually$ and $\until$) has a rational expiration probability $\theta$ in range $(0, 1)$.
For example, $ \eventually_{0.9} \textit{goal} \land \always_{0.9} \textit{lava}$ is a valid GLTL formula.

The semantics of GLTL is similar to that of LTL (which we review in \Cref{sec:ltl_semantics}), except that each operator expires at every step with the given probability $\theta$ associated with the operator.
In particular, for the expiring operator $\always_\theta \phi$, if $\phi$ is always true prior to an expiration event, then the overall formula evaluates to true; otherwise, $\phi$ is ever false prior to the expiration event, then the overall formula evaluates to false.
Conversely, for the expiring operator $\eventually_\theta \phi$, if $\phi$ is ever true prior to an expiration event, then the overall formula evaluates to true; otherwise, $\phi$ is always false prior to the expiration event, then the overall formula evaluates to false.

We give the formal semantics of GLTL below.

We first define the {\em event form} of a GLTL formula. 
An event-form of a GLTL formula is an LTL formula.
This LTL formula contains the propositions in the GLTL formula and an additional set of propositions called {\em expiration events}. 
We define the event-form of a GLTL formula in such a way that, when an expiration event triggers at time $t$, the entire sub-formula corresponding to this event expires. 
The event form $\mathcal{T}(\phi)$ of a GLTL formula $\phi$ is defined recursively as:
\newcommand{\interrupttrans}[1]{\mathcal{T}\mathopen{}\left(#1\right)\mathclose{}}
\begin{equation}
\label{eq:gnl_event_interrupttrans}
\mathcal{T}(\phi) \triangleq
\begin{cases}
\neg \interrupttrans{\psi} & \phi = \neg \psi \\
\interrupttrans{\psi_1} \land \interrupttrans{\psi_2} & \phi = \psi_1 \land \psi_2 \\ 
\interrupttrans{\psi_1} \lor \interrupttrans{\psi_2} & \phi = \psi_1 \lor \psi_2 \\ 
\ltlnext \interrupttrans{\psi} & \phi = \ltlnext \psi \\
\interrupttrans{\psi} \until \left(\interrupttrans{\psi} \land e_\phi\right)
 & \phi = \always_{\theta} \psi \\
 \neg e_\phi \until \interrupttrans{\psi}
 & \phi = \eventually_{\theta} \psi \\
\left(\interrupttrans{\psi_1} \land \neg e_\phi\right) 
\until \interrupttrans{\psi_2} 
 & \phi = \psi_1 \until_{\theta} \psi_2 \\
\end{cases}
\end{equation}
Here, each $e_\phi$ is a fresh proposition corresponding to the sub-formula $\phi$. 
In words:
\begin{itemize}
\item If the outer formula is logical connective or temporal next, the operator $\mathcal{T}$ simply recursively converts the sub-formula(s) while preserving the outer formula's operator.
\item If the outer formula is $\always_\theta$, it recursively converts the sub-formula. 
It outputs the LTL formula $\interrupttrans{\psi} \until \left(\interrupttrans{\psi_2} \land e_\phi\right)$
that requires the converted sub-formula to hold until the expiration event $e_\phi$ expires.
\item If the outer formula is $\eventually_\theta$, it recursively converts the sub-formula. 
It outputs the LTL formula $\neg \interrupttrans{\psi} \until e_\phi$
that requires the converted sub-formula to become true at least once before the expiration event expires.
\item If the outer formula is $\psi_1 \until_\theta \psi_2$, it recursively converts the sub-formula and outputs the LTL formula $\left(\interrupttrans{\psi_1} \land \neg e_\phi\right) 
\until \interrupttrans{\psi_2}$
that requires the converted sub-formula $\mathcal{T}(\psi_1)$ to hold until $\mathcal{T}(\psi_2)$ becomes true, all before the expiration event expires.
\end{itemize}

For example, the event form of the GLTL formula $\phi = \always_{0.9} (a \land \eventually_{0.9} b)$ 
is $$\interrupttrans{\phi} = \left(a \land \left(\neg e_\psi \until b\right)\right) \until \left(\left(a \land \left(\neg e_\psi \until b\right)\right) \land e_\phi \right), \text{ where } \psi \text{ is the sub-formula } \eventually_{0.9} b.$$

During an evaluation of a GLTL formula, each expiration event $e_\phi$ corresponds to an infinite stream of independently distributed Bernoulli random variables, each triggering with probability $\theta$, the expiration probability associated with the outermost expiring temporal operator of $\phi$.  
Given an infinite-length path of propositions $w$, the probability of $w$ satisfying a GLTL formula $\phi$ is the probability of $w$ satisfying the event-form LTL formula $\mathcal{T}(\phi)$, with stochasticity due to the expiration events.

\subsection{GLTL Objective}
We now give the objective specified by a GLTL formula.
A GLTL formula $\phi$ over propositions $\Pi$ specifies an environment-generic objective $\functiontype[\denotation{\phi}]{\infseq{(2^\Pi)}}{\reals}$, given by:
\begin{equation}
\label{eq:gltl_satisfaction_probability_function}
\denotation{\phi}(w) = \prob{\bm{e} \sim \text{Bernoulli}(\bm{\theta})}{ \left(\weunion{w}{w_{\bm{e}}} \vDash \mathcal{T}(\phi)  \right)}.
\end{equation}
Here, $\bm{e}$ is the set of all the expiration events in $\phi$, and each is a random variable following an infinite stream of Bernoulli distribution. 
We use $\weunion{w}{w_{\bm{e}}}: \infseq{(2^{\Pi \cup \bm{e}})}$ to denote the element-wise combination of $w$ and the $w_{\bm{e}}$.
We write $\weunion{w}{w_{\bm{e}}} \vDash \mathcal{T}(\phi)$ to denote that the infinite-length path $\weunion{w}{w_{\bm{e}}}$ satisfies the LTL formula $\mathcal{T}(\phi)$ according to the LTL semantics, which we review in \Cref{sec:ltl_semantics}.
The objective is then the probability of this infinite-length path satisfying the LTL formula $\mathcal{T}(\phi)$, with stochasticity due to the expiration events.

\subsection{Proof of PAC-learnability}

We now prove that the objective specified by a GLTL formula is PAC-learnable.

\begin{proposition}
The objective $\denotation{\phi}$ specified by a GLTL formula $\phi$ is $\denotation{\phi}$-PAC-learnable.
\end{proposition}

\paragraph{Proof Outline} 

The strategy of our proof is as follows.
\begin{itemize}
\item First, we will prove that a GLTL objective is uniformly continuous. In particular, for any $\epsilon \in \rationals$, we will give an upper bound $H$, so that the objective maps infinite-length paths inputs sharing a prefix of $H$ to $\epsilon$-close values.
\item Then, we give a pseudocode that computes a GLTL objective. In particular, to give the $n$-th rational approximation, the pseudocode first computes the upper bound $H$ from $\epsilon = 2^{-n}$. It then computes a lower bound of the objective value after observing the first $H$ indices of the input. It returns the lower bound value as the $n$-th rational approximation.
\item Finally, we show that the rational approximations form a fast-converging Cauchy sequence, which proves that the objective is computable. Therefore by \Cref{thm:computable-implies-uc}, the objective is PAC-learnable.
\end{itemize}

\paragraph{GLTL Objective is Uniformly Continuous}
First, we give the following \Cref{lemma:simu_expire} that if all expiration events simultaneously trigger at step $H$, then the evaluation of the event-form of the GLTL formula only depends on the length-$H$ prefix of the infinite-length path input. 
\begin{lemma}[Simultaneous Expiration]
\label{lemma:simu_expire}
Given a GLTL formula $\phi$, the satisfaction of its event-form depends on the input infinite-length path up to a horizon that all expiration events are simultaneously triggered. That is:
\begin{equation}
\begin{split}
\forall w_{\bm{e}} \in \infseq{(2^{\bm{e}})}. \forall w_1 \in \infseq{(2^\Pi)}. \forall w_2 \in \infseq{(2^\Pi)} \colon   \\
\text{if } \left(\windex{w_{\bm{e}}}{H} = \vec{1} \land
\wprefix{w_1}{H} = \wprefix{w_2}{H} \right)
\text{then } \Big( \weunion{w_1}{w_{\bm{e}}} \vDash \interrupttrans{\phi} \iff \weunion{w_2}{w_{\bm{e}}} \vDash \interrupttrans{\phi}\Big)
\end{split}
\end{equation}
\end{lemma}
Then, we utilize \Cref{lemma:simu_expire} to prove the following lemma that a GLTL objective is uniformly continuous.
\begin{lemma}
\label{lemma:gltl-is-uc}
A GLTL objective $\denotation{\phi}$ is uniformly continuous.
\end{lemma}
\begin{proof}
For any input word $w$, we rewrite the value of the objective in \Cref{eq:gltl_satisfaction_probability_function} by unrolling the first $H$ steps.
Then, we condition the objective (which is a probability) on if all expirations expire simultaneously at each step.
In particular, let $E_k$ denote the event that all expirations trigger simultaneously at step $k$ and let $\neg E_{1\dots H}$ denote the event that all expiration never simultaneously trigger before step $H$. 
We expand \Cref{eq:gltl_satisfaction_probability_function} as:
\begin{equation}
\label{eq:gltl_satisfaction_probability_function-expand}
\begin{split}
\denotation{\phi}(w) &= \sum_{k=1}^H \prob{}{E_k} \cdot \prob{}{ \weunion{w}{w_{\bm{e}}} \vDash \interrupttrans{\phi} \mid E_k}+ \\ 
& \prob{}{\neg E_{1 \dots H}} \cdot \prob{}{\weunion{w}{w_{\bm{e}}} \vDash \interrupttrans{\phi}  | \neg E_{1 \dots H}} 
\end{split}
\end{equation}
Then, consider two paths $w_1$ and $w_2$ sharing a prefix of length $H$.
By \Cref{lemma:simu_expire},  the satisfaction of $\interrupttrans{\phi}$ depends only on the first $H$ steps of $w_1$ and $w_2$. 
Thus, for all $0 \le k \le H$, $\prob{}{ \weunion{w_1}{w_{\bm{e}}} \vDash \interrupttrans{\phi} \mid E_k} = \prob{}{ \weunion{w_2}{w_{\bm{e}}} \vDash \interrupttrans{\phi} \mid E_k}$.
Therefore, for the difference $\Delta = |\denotation{\phi}(w_1) - \denotation{\phi}(w_2)|$, we cancel the first $H$ terms in the sum to get:
\begin{equation*}
\begin{split}
\Delta &= \prob{}{\neg E_{1 \dots H}} \cdot \left| \prob{}{\weunion{w_1}{w_{\bm{e}}} \vDash \interrupttrans{\phi}  | \neg E_{1 \dots H}} - \prob{}{\weunion{w_2}{w_{\bm{e}}} \vDash \interrupttrans{\phi}  | \neg E_{1 \dots H}} \right| \\ 
& \le \prob{}{\neg E_{1 \dots H}}
\end{split}
\end{equation*}
Since $\prob{}{\neg E_{1 \dots H}}$ is the probability of the all the expirations not triggering simultaneously for the first $H$ steps, we have:
$\prob{}{\neg E_{1 \dots H}} = (1 - \prod_{\theta_i \in \phi} \theta_i )^{H}$.
Consequently, we obtain the upper bound $\Delta \le (1 - \prod_{\theta_i \in \phi} \theta_i )^{H}$.

Given any $\epsilon > 0$, we can always choose $H = \frac{\log(\epsilon)}{\log(1 - \prod_{\theta_i \in \phi} \theta_i)}$ so that $\Delta \le \epsilon$.
Therefore, $\denotation{\phi}$ is uniformly continuous.
\end{proof}

\begin{pseudocode}[label={lst:gltl-objective}, caption={Computation of a GLTL objective}]
# Given GLTL formula (*$\phi$*)
def GLTL(w: (*$\infseq{(2^\Pi)}$*), n: (*$\naturals$*)) -> (*$\rationals$*):
  theta = prod((*$\phi$*).thetas) # take product of all the expiration probabilities in (*$\phi$*)
  H = -n / log2ceil(1 - theta)
  v = 0
  for w_e in (*$\bm{e}^\text{H}$*): 
    if w_e has a simultaneously trigger:
      w' = (*$\wrational{\weunion{w[:H]}{\text{w\_e}}}{\cdot}$*)
      if (*$w' \vDash \mathcal{T}(\phi)$*): 
        e_prob = 1
        for e_i in (*$\bm{e}$*): # enumerates all the events in the formula
          w_e_i = w_e[e_i]
          for j in range(H): # enumerates the values in the length-H stream of an event
            e_prob *= w_e_i.theta if w_e_i[j] == 1 else (1 - w_e_i.theta) # event triggers with probability w_e_i's theta
        v += e_prob
  return v
\end{pseudocode}

\paragraph{Computation of a GLTL Objective}

We give pseudocode for computing a GLTL objective in \Cref{lst:gltl-objective}.
The code in \Cref{lst:gltl-objective} first computes a horizon $H = \frac{-n}{\lceil\log_2(1 - \prod_{\theta_i \in \phi} \theta_i)\rceil}$. 
It then computes the sum of $H$-terms in \Cref{eq:gltl_satisfaction_probability_function-expand}:
\begin{equation}
\label{eq:gltl_satisfaction_probability_function-partial-sum}
\sum_{k=1}^H \prob{}{E_k} \cdot \prob{}{ \weunion{w}{w_{\bm{e}}} \vDash \interrupttrans{\phi} \mid E_k} .
\end{equation}
To compute this sum, it enumerates all combinations of expiration events of length $H$.
If a simultaneous expiration ever happens in this length $H$ events, it tests if the input word is accepted or rejected by $\mathcal{T}(\phi)$. 
In particular, to perform the test, it forms any eventual cyclic path $w' = \wrational{\weunion{w[:H]}{w_{\bm{e}}}}{\cdot}$ that starts with the first $H$ indices of the input $w$ and the length-$H$ sequence of events $w_{\bm{e}}$ and ends in an arbitrarily chosen cycle. 
It then test if $w'$ satisfies $\mathcal{T}(\phi)$ by a standard LTL model checking algorithm \citep[Chapter~5.2]{principlesofmodelchecking}.
Due to \Cref{lemma:simu_expire}, this test is equivalent to testing if $\weunion{w}{w_{\bm{e}}}$ satisfies $\mathcal{T}(\phi)$. 
Finally, it sums the probability of all length-$H$ sequence of expiration events that pass this test.
This produces the desired sum in \Cref{eq:gltl_satisfaction_probability_function-partial-sum}.

\paragraph{Conclusion}

By the proof of \Cref{lemma:gltl-is-uc}, the sum in \Cref{eq:gltl_satisfaction_probability_function-partial-sum} is at most $\epsilon = 2^{-n}$ smaller than the objective's value.
Therefore, the return values of \Cref{lst:gltl-objective} for all $n\in\naturals$ form a fast-converging Cauchy sequence that converges to the objective's value. 
Thus, the objective is computable and $\denotation{\phi}$-PAC-learnable.

\section{Proof of \Cref{lemma:simu_expire}}
\label{sec:proof_simu_expire}

We perform our proof inductively. Specifically, given any GLTL formula $\phi$, we pose the inductive hypothesis that each sub-formula of $\phi$ satisfies \Cref{lemma:simu_expire}. 
Given this inductive hypothesis, we prove that $\phi$ also satisfies \Cref{lemma:simu_expire}.

\paragraph{If the outer formula of $\phi$ is $\neg \psi$}

Suppose that $w_{\bm{e}}$ is true at step $H$ and that $w_1$ and $w_2$ match up to step $H$:
$$\windex{w_{\bm{e}}}{H} = \vec{1} \land
\wprefix{w_1}{H} = \wprefix{w_2}{H} .$$
By the induction hypothesis, the evaluations of the sub-formula $\psi$ are the same for the two inputs $w_1$ and $w_2$: $$\weunion{w_1}{w_{\bm{e}}} \vDash \interrupttrans{\psi} \iff \weunion{w_2}{w_{\bm{e}}} \vDash \interrupttrans{\psi}.$$
Therefore, we can prepend negations to both sides: $$\weunion{w_1}{w_{\bm{e}}} \vDash \neg \interrupttrans{\psi} \iff \weunion{w_2}{w_{\bm{e}}} \vDash \neg \interrupttrans{\psi} .$$
By definition of $\mathcal{T}$, we have:
$\weunion{w_1}{w_{\bm{e}}} \vDash \interrupttrans{\phi} \iff \weunion{w_2}{w_{\bm{e}}} \vDash \interrupttrans{\phi}$, which proves this case.

\paragraph{If the outer formula of $\phi$ is $\psi_1 \land \psi_2$ or $\psi_1 \lor \psi_2$}

We will prove for the case of $\phi = \psi_1 \land \psi_2$. The case of $\phi = \psi_1 \lor \psi_2$ is essentially the same by changing the logical connective from $\land$ to $\lor$.

Suppose that $w_{\bm{e}}$ is true at step $H$ and that $w_1$ and $w_2$ match up to step $H$:
$$\windex{w_{\bm{e}}}{H} = \vec{1} \land
\wprefix{w_1}{H} = \wprefix{w_2}{H} .$$
By the induction hypothesis, the evaluations of each of the sub-formula $\psi_1$ and $\psi_2$ are the same for the two inputs $w_1$ and $w_2$, that is: $$\weunion{w_1}{w_{\bm{e}}} \vDash \interrupttrans{\psi_1} \iff \weunion{w_2}{w_{\bm{e}}} \vDash \interrupttrans{\psi_1} \text{ and } \weunion{w_1}{w_{\bm{e}}} \vDash \interrupttrans{\psi_2} \iff \weunion{w_2}{w_{\bm{e}}} \vDash \interrupttrans{\psi_2} .$$
Therefore, we can join the two equivalences by the logical connective: $\weunion{w_1}{w_{\bm{e}}} \vDash \interrupttrans{\psi_1} \land \interrupttrans{\psi_2} \iff \weunion{w_2}{w_{\bm{e}}} \vDash \interrupttrans{\psi_1} \land \interrupttrans{\psi_2}$. 
By definition of $\mathcal{T}$, we have:
$\weunion{w_1}{w_{\bm{e}}} \vDash \interrupttrans{\phi} \iff \weunion{w_2}{w_{\bm{e}}} \vDash \interrupttrans{\phi}$, which proves this case.

\paragraph{If the outer formula of $\phi$ is $\always_\theta \psi$}

Suppose that $w_{\bm{e}}$ is true at step $H$ and that $w_1$ and $w_2$ match up to step $H$:
$$\windex{w_{\bm{e}}}{H} = \vec{1} \land
\wprefix{w_1}{H} = \wprefix{w_2}{H} .$$
Consider the infinite-length paths $w_1^i \triangleq \wsuffix{w_1}{i}$, $w_2^i \triangleq \wsuffix{w_2}{i}$ and $w_{\bm{e}}^i \triangleq \wsuffix{w_{\bm{e}}}{i}$, the suffixes of $w_1$, $w_2$ and $w_{\bm{e}}$ beginning at some step $i$ where $0 \le i \le H$. 
The expiration event triggers at step $H-i$ for  $\wsuffix{w_{\bm{e}}}{i}$, that is: $\windex{w_{\bm{e}}^i}{H-i} = \vec{1}$.
Further, $w_1^i$ and $w_2^i$ match up to length $H-i$, that is: $ \wsuffix{w_1^i}{H - i} = \wsuffix{w_2^i}{H - i}$.
Therefore, by the induction hypothesis, the evaluations of the sub-formula $\psi$ are the same for all pairs of suffix inputs $w_1^i$ and $w_2^i$ for all $i$, that is: 
\begin{equation}
\label{eq:gltl_lemma_g_eq1}
\forall 0 \le i \le H. \wsuffix{\weunion{w_1}{w_{\bm{e}}}}{i} \vDash \interrupttrans{\psi} \iff \wsuffix{\weunion{w_2}{w_{\bm{e}}}}{i} \vDash \interrupttrans{\psi}.
\end{equation}
Since $w_{\bm{e}}$ is fixed on both sides, conjuncting both sides with $e_\phi$, we also have:
\begin{equation}
\label{eq:gltl_lemma_g_eq2}
\forall 0 \le i \le H. \wsuffix{\weunion{w_1}{w_{\bm{e}}}}{i} \vDash \left(\interrupttrans{\psi} \land e_\phi\right) \iff \wsuffix{\weunion{w_2}{w_{\bm{e}}}}{i} \vDash \left(\interrupttrans{\psi} \land e_\phi \right).
\end{equation}
To summarize \Cref{eq:gltl_lemma_g_eq1,eq:gltl_lemma_g_eq2}: The satisfaction relations in the above equations are equal between $w_1$ and $w_2$ up to step $H$. 

By defintion of $\mathcal{T}$ and the semantics of LTL (reviewed in \Cref{sec:ltl_semantics}), $\weunion{w}{w_{\bm{e}}} \vDash \interrupttrans{\phi}$ iff:
\begin{equation}
\label{eq:gltl_lemma_g_eq3}
\exists j \ge 0, \wsuffix{\weunion{w}{w_{\bm{e}}}}{j} \vDash (\interrupttrans{\psi} \land e_\phi) \text{ and }  \forall k\ldotp 0 \le k < j \implies \wsuffix{\weunion{w}{w_{\bm{e}}}}{k} \vDash \interrupttrans{\psi}. 
\end{equation}

If $\wsuffix{\weunion{w_1}{w_{\bm{e}}}}{j} \vDash (\interrupttrans{\psi} \land e_\phi)$ and $\wsuffix{\weunion{w_1}{w_{\bm{e}}}}{j} \vDash (\interrupttrans{\psi} \land e_\phi)$ are both true for any $j \le H$, then $\weunion{w_1}{w_{\bm{e}}} \vDash \interrupttrans{\phi}$ equals $\weunion{w_2}{w_{\bm{e}}} \vDash \interrupttrans{\phi}$ due to all satisfaction relations in \Cref{eq:gltl_lemma_g_eq3} match between $w_1$ and $w_2$ up to step $H$.

On the other hand, if $\wsuffix{\weunion{w_1}{w_{\bm{e}}}}{j} \vDash (\interrupttrans{\psi} \land e_\phi)$ and $\wsuffix{\weunion{w_1}{w_{\bm{e}}}}{j} \vDash (\interrupttrans{\psi} \land e_\phi)$ are never true for some $j \le H$,  $\weunion{w_1}{w_{\bm{e}}} \vDash \interrupttrans{\phi}$ and$\weunion{w_2}{w_{\bm{e}}} \vDash \interrupttrans{\phi}$ both equal false. 
That is because, since $e_\phi$ is true at step $H$, in order for $\wsuffix{\weunion{w}{w_{\bm{e}}}}{H} \vDash (\interrupttrans{\psi} \land e_\phi)$ to be false, $\mathcal{T}(\psi)$ must be false at step $H$ --- which in turn implies that there is no $j$ such that $\mathcal{T}(\psi)$ will hold until $\mathcal{T}(\psi) \land e_\phi$ becomes true at step $j$.

In both scenarios, we have $\weunion{w_1}{w_{\bm{e}}} \vDash \interrupttrans{\phi}$ equivalent to $ \weunion{w_2}{w_{\bm{e}}} \vDash \interrupttrans{\phi}$, which proves this case.

\paragraph{If the outer formula of $\phi$ is $\psi_1 \until_\theta \psi_2$}

Suppose that $w_{\bm{e}}$ is true at step $H$ and that $w_1$ and $w_2$ match up to step $H$:
$$\windex{w_{\bm{e}}}{H} = \vec{1} \land
\wprefix{w_1}{H} = \wprefix{w_2}{H} .$$
Consider the infinite-length paths $w_1^i = \wsuffix{w_1}{i}$, $w_2^i = \wsuffix{w_2}{i}$ and $w_{\bm{e}}^i = \wsuffix{w_{\bm{e}}}{i}$, the suffixes of $w_1$, $w_2$ and $w_{\bm{e}}$ beginning at some step $i$ where $0 \le i \le H$. 
The expiration event triggers at step $H-i$ for  $\wsuffix{w_{\bm{e}}}{i}$, that is: $\windex{w_{\bm{e}}^i}{H-i} = \vec{1}$.
Further, $w_1^i$ and $w_2^i$ match up to length $H-i$, that is: $ \wsuffix{w_1^i}{H - i} = \wsuffix{w_2^i}{H - i}$.
Therefore, by the induction hypothesis, the evaluations of each of the sub-formulas $\psi_1$ and $\psi_2$ are the same for all pairs of suffix inputs $w_1^i$ and $w_2^i$ for all $i$, that is: 
\begin{equation}
\label{eq:gltl_lemma_u_eq1}
\begin{split}
\forall 0 \le i \le H. \wsuffix{\weunion{w_1}{w_{\bm{e}}}}{i} \vDash \interrupttrans{\psi_1} \iff \wsuffix{\weunion{w_2}{w_{\bm{e}}}}{i} \vDash \interrupttrans{\psi_1} \text{ and }\\
\forall 0 \le i \le H. \wsuffix{\weunion{w_1}{w_{\bm{e}}}}{i} \vDash \interrupttrans{\psi_2} \iff \wsuffix{\weunion{w_2}{w_{\bm{e}}}}{i} \vDash \interrupttrans{\psi_2}.
\end{split}
\end{equation}
Since $w_{\bm{e}}$ is fixed on both sides, conjuncting both sides with $e_\phi$, we also have:
\begin{equation}
\label{eq:gltl_lemma_u_eq2}
\begin{split}
\forall 0 \le i \le H. \wsuffix{\weunion{w_1}{w_{\bm{e}}}}{i} \vDash \left(\interrupttrans{\psi_1} \land e_\phi\right) \iff \wsuffix{\weunion{w_2}{w_{\bm{e}}}}{i} \vDash \left(\interrupttrans{\psi_1} \land e_\phi \right). \text{ and }\\
\forall 0 \le i \le H. \wsuffix{\weunion{w_1}{w_{\bm{e}}}}{i} \vDash \left(\interrupttrans{\psi_2} \land e_\phi\right) \iff \wsuffix{\weunion{w_2}{w_{\bm{e}}}}{i} \vDash \left(\interrupttrans{\psi_2} \land e_\phi \right).
\end{split}
\end{equation}
To summarize \Cref{eq:gltl_lemma_u_eq1,eq:gltl_lemma_u_eq2}: The satisfaction relations in the above equations are equal between $w_1$ and $w_2$ up to step $H$. 

By definition of $\mathcal{T}$ and the semantics of LTL (reviewed in \Cref{sec:ltl_semantics}), $\weunion{w}{w_{\bm{e}}} \vDash \interrupttrans{\phi}$ iff:
\begin{equation}
\label{eq:gltl_lemma_u_eq3}
\exists j \ge 0, \wsuffix{\weunion{w}{w_{\bm{e}}}}{j} \vDash \interrupttrans{\psi_2} \text{ and }  \forall k\ldotp 0 \le k < j \implies \wsuffix{\weunion{w}{w_{\bm{e}}}}{k} \vDash \left(\interrupttrans{\psi_1} \land \neg e_\phi\right). 
\end{equation}

If $\wsuffix{\weunion{w_1}{w_{\bm{e}}}}{j} \vDash \interrupttrans{\psi_2}$ and $\wsuffix{\weunion{w_1}{w_{\bm{e}}}}{j} \vDash \interrupttrans{\psi_2}$ are both true for any $j \le H$, then $\weunion{w_1}{w_{\bm{e}}} \vDash \interrupttrans{\phi}$ equals $\weunion{w_2}{w_{\bm{e}}} \vDash \interrupttrans{\phi}$ due to all satisfaction relations in \Cref{eq:gltl_lemma_u_eq3} match between $w_1$ and $w_2$ up to step $H$.

On the other hand, if $\wsuffix{\weunion{w_1}{w_{\bm{e}}}}{j} \vDash \interrupttrans{\psi_2}$ and $\wsuffix{\weunion{w_1}{w_{\bm{e}}}}{j} \vDash \interrupttrans{\psi_2}$ are never true for some $j \le H$,  $\weunion{w_1}{w_{\bm{e}}} \vDash \interrupttrans{\phi}$ and$\weunion{w_2}{w_{\bm{e}}} \vDash \interrupttrans{\phi}$ both equal false. 
That is because, since $e_\phi$ is true at step $H$, $\interrupttrans{\psi_2} \land \neg e_\phi$ is false at step $H$ --- which in turn implies that there is no $j$ such that $\mathcal{T}(\psi_1) \land \neg e_\phi$ will hold until $\mathcal{T}(\psi_2)$ becomes true at step $j$.

In both scenarios, we have $\weunion{w_1}{w_{\bm{e}}} \vDash \interrupttrans{\phi}$ equivalent to $ \weunion{w_2}{w_{\bm{e}}} \vDash \interrupttrans{\phi}$, which proves this case.

\paragraph{If the outer formula of $\phi$ is $\eventually_\theta \psi$}
By definition of the conversion operator $\mathcal{T}$, it is the case that $\interrupttrans{\eventually_\theta \psi} = \interrupttrans{\texttt{true} \until_\theta \psi}$. 
Therefore the proof of the case of $\phi = \eventually_\theta \psi$ is the same as the proof of the case of $\phi = \psi_1 \until_\theta \psi_2$, by specializing $\psi_1 = \texttt{true}$ and $\psi_2 = \psi$.

\section{LTL Semantics}
\label{sec:ltl_semantics}

This section reviews linear temporal logic (LTL) and its semantics. 
For a more in-depth introduction of LTL, we refer readers to  \citet{principlesofmodelchecking}.

An LTL formula is built from a finite set of atomic propositions $\Pi$, logical connectives, temporal next $\ltlnext$, and temporal operators  $\always$ (always), $\eventually$ (eventually), and $\until$ (until). 
\Cref{eq:ltl_grammar} gives the grammar of an LTL formula $\phi$ over the set of atomic propositions $\Pi$:
\Cref{eq:gnl_grammar} gives the grammar of an LTL formula $\phi$ over the set of atomic propositions $\Pi$:
\begingroup
\newcommand{\alt}{\:\big|\:}%
\begin{equation}
\label{eq:ltl_grammar}
\phi \defeq 
a
\alt \neg \phi 
\alt \phi \land \phi 
\alt \phi \lor \phi
\alt \ltlnext \phi
\alt \always \phi 
\alt \eventually \phi
\alt \phi \until \phi, \; a \in \Pi
\end{equation}
\endgroup

LTL is a logic over infinite-length paths.
For an LTL formula $\phi$, we write $w \vDash \phi$ to denote that the infinite-length path $w$ satisfies $\phi$.
The following \Cref{eq:ltl_semantics} fully defines this satisfaction relation:
\begin{definition}[LTL Semantics]
\begin{equation}
\label{eq:ltl_semantics}
\newcommand{\txtiff}{&\text{ iff }&}
\begin{aligned}
w &\vDash a  \txtiff a \in \windex{w}{i} \quad a \in \Pi \\
w &\vDash \neg \phi \txtiff w \nvDash \phi \\
w &\vDash \phi \land \psi \txtiff w \vDash \phi \text{ and } w \vDash \psi \\
w &\vDash \phi \lor \psi \txtiff w \vDash \phi \text{ or } w \vDash \psi \\
w &\vDash \ltlnext \phi \txtiff \wsuffix{w}{1} \vDash \phi \\
w &\vDash \always \phi \txtiff \forall i \ge 0, \wsuffix{w}{i} \vDash \phi \\
w &\vDash \eventually \phi \txtiff \exists i \ge 0, \wsuffix{w}{i} \vDash \phi \\
w &\vDash \phi \until \psi \txtiff \exists j \ge 0, \wsuffix{w}{j} \vDash \psi \text{ and }  \forall k\ldotp 0 \le k < j \implies \wsuffix{w}{k} \vDash \phi. \\
\end{aligned}
\end{equation}
\end{definition}
In words, the semantics of each temporal operator is:
\begin{itemize}
\item $\ltlnext \phi$: the sub-formula $\phi$ is true in the next time step.
\item $\always \phi$: the sub-formula $\phi$ is always true in all future time steps.
\item $\eventually \phi$: the sub-formula $\phi$ is eventually true in some future time steps.
\item $\phi \until \psi$: the sub-formula $\phi$ is always true until the sub-formula $\psi$ eventually becomes true, after which $\phi$ is allowed to become false.
\end{itemize}

\end{appendices}
\fi

\end{document}